\documentclass[final,3p,times]{elsarticle}

\RequirePackage{dsfont}

\def\sign{sign}

\usepackage{amssymb,amsmath}
\usepackage{graphicx}
\usepackage{color}
\usepackage{subfigure}
\graphicspath{{images/}}
\usepackage{multirow}
\usepackage{float}

\journal{Information Fusion}

\begin{document}

\begin{frontmatter}



\title{Features modeling with an $\alpha$-stable distribution: application to pattern recognition based on continuous belief functions}

\author[label2]{Anthony Fiche\corref{cor1}}
\ead{anthony.fiche@ensta-bretagne.fr}
\author[label2]{Jean-Christophe Cexus}
\ead{jean-christophe.cexus@ensta-bretagne.fr}
\author[label3]{Arnaud Martin}
\ead{arnaud.martin@univ-rennes1.fr}
\author[label2]{Ali Khenchaf}
\ead{ali.khenchaf@ensta-bretagne.fr}
\cortext[cor1]{Corresponding author.}
\address[label2]{LabSticc, UMR 6285, ENSTA Bretagne, 2 rue Fran\c cois Verny, 29806 Brest Cedex 9, France}
\address[label3]{UMR 6074 IRISA, IUT Lannion / Universit\'e de Rennes 1, rue \'Edouard Branly BP 30219, 22302 Lannion Cedex, France} 

\begin{abstract}
The aim of this paper is to show the interest in fitting features with an $\alpha$-stable distribution to classify imperfect data. The supervised pattern recognition is thus based on the theory of continuous belief functions, which is a way to consider imprecision and uncertainty of data. The distributions of features are supposed to be unimodal and estimated by a single Gaussian and $\alpha$-stable model. Experimental results are first obtained from synthetic data by combining two features of one dimension and by considering a vector of two features. Mass functions are calculated from plausibility functions by using the generalized Bayes theorem. The same study is applied to the automatic classification of three types of sea floor (rock, silt and sand) with features acquired by a mono-beam echo-sounder. We evaluate the quality of the $\alpha$-stable model and the Gaussian model by analyzing qualitative results, using a Kolmogorov-Smirnov test (K-S test), and quantitative results with classification rates. The performances of the belief classifier are compared with a Bayesian approach. 

\end{abstract}


\end{frontmatter}


\section{Introduction}
The choice of a model has an important role in the problem of estimation. For example, the Gaussian model is a very efficient model which fits data in many applications as it is very simple to use and saves computation time. However, as is the case for all distribution models, Gaussian laws have some weaknesses and results can end-up being skewed. Indeed, as the Gaussian probability density function ({\it pdf}) is symmetrical, it is not valid when the {\it pdf} is not symmetrical. It is therefore difficult to choose the right model which fits data for each application. The Gaussian distribution belongs to a family of distributions called stable distributions. This family of distributions allows the representation of heavy tails and skewness. A distribution is said to have a heavy tail if the tail decays slower than the tail of the Gaussian distribution. Therefore, the property of skewness means that it is impossible to find a mode where probability density function is symmetrical. The main property of stable laws introduced by L\'evy~\cite{Lévy1924} is that the sum of two independent stable random variables gives a stable random variable. $\alpha$-stable distributions are used in different fields of research such as radar~\cite{achim2003sar,banerjee1999adaptive}, image processing~\cite{kuruoglu2003skewed} or finance~\cite{mcculloch1996financial,fama1965behavior}, ...

The aim of this study is to show the interest in fitting data with an $\alpha$-stable distribution. In~\cite{williams}, the author proposes to characterize the sea floor from a vector of features modeled by Gaussian mixture models (GMMs) using an Autonomous Underwater Vehicle (AUV). However, the problem with the Bayesian approach is the difficulty in considering the uncertainty of data. We therefore favored the use of an approach based on the theory of belief functions~\cite{Dempster1967,Shafer1976}. In~\cite{fiche2009}, the authors compared a Bayesian and belief classifier where data from sensors are modeled using GMMs estimated \emph{via} an Expectation-Maximization (EM) algorithm~\cite{Dempster1977}. This work has been extended to data modeled by $\alpha$-stable mixture models~\cite{fiche2010}. However, it is difficult to choose between these two models because the results are roughly the same. This paper raises two problems. Firstly, it is necessary to work on a data set where features are modeled by $\alpha$-stable {\it pdf}s. The second problem is to know how to apply the theory of belief functions when data from sensors are modeled by $\alpha$-stable distributions. This point has been dealt with in~\cite{fusion}.

The recent characterization of uncertain environments has taken an important place in several fields of application such as the SOund Navigation And Ranging (SONAR). SONAR has therefore been used for the detection of underwater mines~\cite{quidu}. In~\cite{laanaya}, the author developed techniques to perform automatic classification of sediments on the seabed from sonar images. In~\cite{fiche2009}, the authors classified sonar images by extracting features and modeling them with GMMs. In~\cite{Lurton}, the authors characterized the sea floor using data from a mono-beam echo-sounder. A data set represents an echo signal amplitude according to time. They compare the time envelope of echo signal amplitude with a set of theoretical reference curves. In this paper, we build a classifier based on the theory of belief functions where a vector of features extracted from a mono-beam echo-sounder is modeled by a Gaussian and an $\alpha$-stable distribution. The feature {\it pdf}s have only one mode. We finally show that it would be interesting to fit features with an $\alpha$-stable model.

This paper is organized as follows:  we first introduce the definition of stability, the method to construct $\alpha$-stable probability density functions and the methods of estimation in Section~\ref{section 2}. The definitions in the multivariate case are also presented. We introduce the notion of belief functions in discrete and continuous cases in Section~\ref{section 3}. Belief functions are then calculated in the particular case of $\alpha$-stable distributions and a belief classifier is constructed in Section~\ref{section 4}. We finally classify data by modeling data with Gaussian and $\alpha$-stable distributions before the generated data are classified by comparing the theory of belief functions with Gaussian and $\alpha$-stable distributions in Section~\ref{section 5}. We compare the results obtained with the theory of belief functions using a Bayesian approach.

\section{The $\alpha$-stable distributions}
\label{section 2}
Gauss introduced the probability density function called ``Gaussian distribution'' (in honor of Gauss) in his study of astronomy. Laplace and Poisson developed the theory of characteristic function by calculating the analytical expression of Fourier transform of a probability density function. Laplace found that the Fourier transform of a Gaussian law is also a Gaussian law. Cauchy tried to calculate the Fourier transform of a ``generalized Gaussian'' function with the expression $f_n(x)=\frac{1}{\pi}\int_{0}^{+\infty}\exp{(-ct^n)}\cos{(tx)}dt$, but he did not solve the problem. When the integer $n$ is a real $\alpha$, we define the family of $\alpha$-stable distributions. However, Cauchy did not know at that time if he had defined a probability density function. With the results of Polya and Bernstein, Janicki and Weron~\cite{janicki} demonstrated that the family of $\alpha$-stable laws are probability density functions. The mathematician Paul L\'evy  studied the central limit theorem and showed, with the constraint of infinite variance, that the limit law is a stable law~\cite{Lévy1924}. Motivated by this property, L\'evy calculated the Fourier transform for all $\alpha$-stable distributions. These stable laws which satisfy the generalized central limit theorem are interesting as they allow the modeling of data as impulsive noise when the Gaussian model is not valid. Another property of these laws is the ability to model heavy tails and skewness. In the following, we define the notions of stability, characteristic function, the way to build a {\it pdf} from characteristic function, the different estimators of $\alpha$-stable distributions and the extension of $\alpha$-stable distribution to the multivariate case.
\subsection{Definition of stability}
Stability is when the sum of two independent random variables which follow $\alpha$ laws, gives an $\alpha$ law. Mathematically, this definition means the following: a random variable $X$ is stable, denoted $X\sim S_{\alpha}(\beta,\gamma,\delta)$, if for all ($a$,$b$)~$\in$~$\mathbb{R}^+~\times~\mathbb{R}^+$, there are $c$ $\in$ $\mathbb{R}^+$ and $d$ $\in$ $\mathbb{R}$ such that:\\ 
\begin{equation}
\displaystyle a X_1 + b X_2 = c X + d,
\label{equation1}
\end{equation}
with $X_1$ and $X_2$ two independent $\alpha$-stable random variables which follow the same distribution as $X$. If Equation~\eqref{equation1} defines the notion of stability, it does not give any indication as to how to parameterize an $\alpha$-stable distribution. We therefore prefer to use the definition given by characteristic function to refer to an $\alpha$-stable distribution.

\subsection{Characteristic function}
Several equivalent definitions have been suggested in the literature to parameterize an $\alpha$-stable distribution from its characteristic function~\cite{taqqu,zolotarev}. Zolotarev~\cite{zolotarev} proposed the following:
\begin{equation}
\phi_{S_{\alpha}(\beta,\gamma,\delta)}(t)= \left \lbrace 
\begin{array}{l c} 
\displaystyle \exp(j t \delta - |\gamma t|^\alpha [1+ j \beta \tan(\frac{\pi \alpha}{2})\sign(t) (|t|^{1-\alpha} - 1)])&\text{ if } \alpha \neq 1,\\
\displaystyle \exp(j t \delta - |\gamma t| [1 + j \beta \frac{2}{\pi} \sign(t) \log|t|])&\text{ if } \alpha = 1,
\end{array}              
\right .
\label{cdf_zolotarev}  
\end{equation}

where each feature has specific values:
\begin{itemize}
\item $\displaystyle\alpha \in]0,2]$ is the characteristic exponent.
\item $\displaystyle\beta \in[-1,1]$ is the skewness parameter. 
\item $\gamma \in \mathbb{R}^{+*}$ represents the scale parameter.
\item $\delta \in \mathbb{R}$ is the location parameter.
\end{itemize}
The advantage of this parameterization compared to \cite{taqqu} is that the values of the characterization and probability density functions are continuous for all parameters. In fact, the parameterization defined by \cite{taqqu} is discontinuous when $\alpha=1$ and $\beta=0$.
\subsection{The probability density function}
The representation of an $\alpha$-stable {\it pdf}, denoted $f_{S_{\alpha}(\beta,\gamma,\delta)}$, is obtained by calculating the Fourier transform of its characteristic function:
\begin{equation}
\displaystyle f_{S_{\alpha}(\beta,\gamma,\delta)}(x)= \int_{-\infty}^{\infty} \phi_{S_{\alpha}(\beta,\gamma,\delta)}(t) \exp(-j t x) dt.   
\label{pdf_zolotarev}
\end{equation}

\begin{figure}[b]
	\centering
		\begin{tabular}{cc}
		
	\subfigure[Influence of parameter $\alpha$ with $\beta=0$, $\gamma=1$ and $\delta=0$.]{\includegraphics[height=6cm,width=7cm]{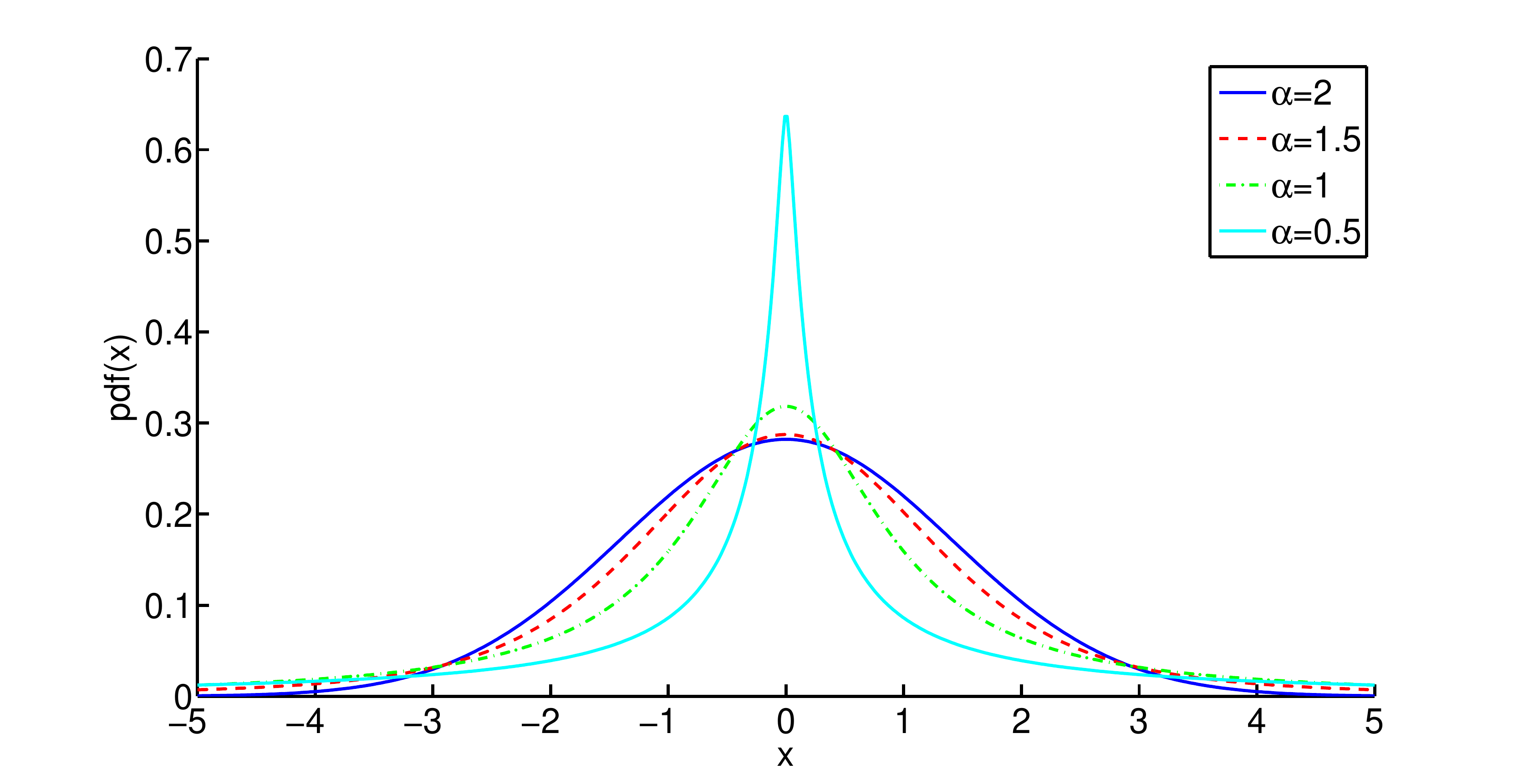}\label{fig:comparaison-alpha}}	&  \subfigure[Influence of parameter $\beta$ with $\alpha=1.5$, $\gamma=1$ and $\delta=0$.]{\includegraphics[height=6cm,width=7cm]{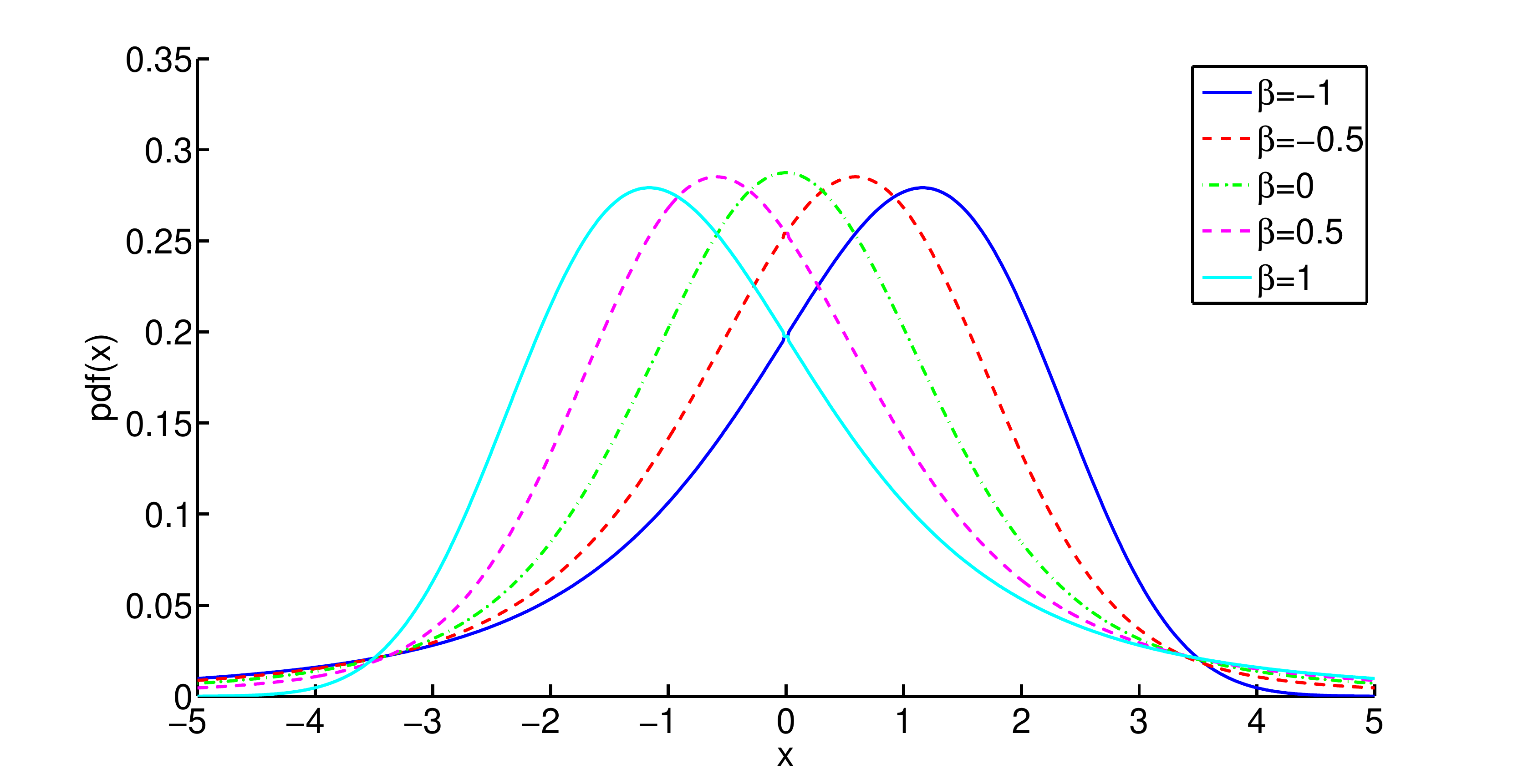}\label{fig:comparaison-beta}}	 \\
\subfigure[Influence of parameter $\gamma$ with $\alpha=1.5$, $\beta=0$ and $\delta=0$.]{\includegraphics[height=6cm,width=7cm]{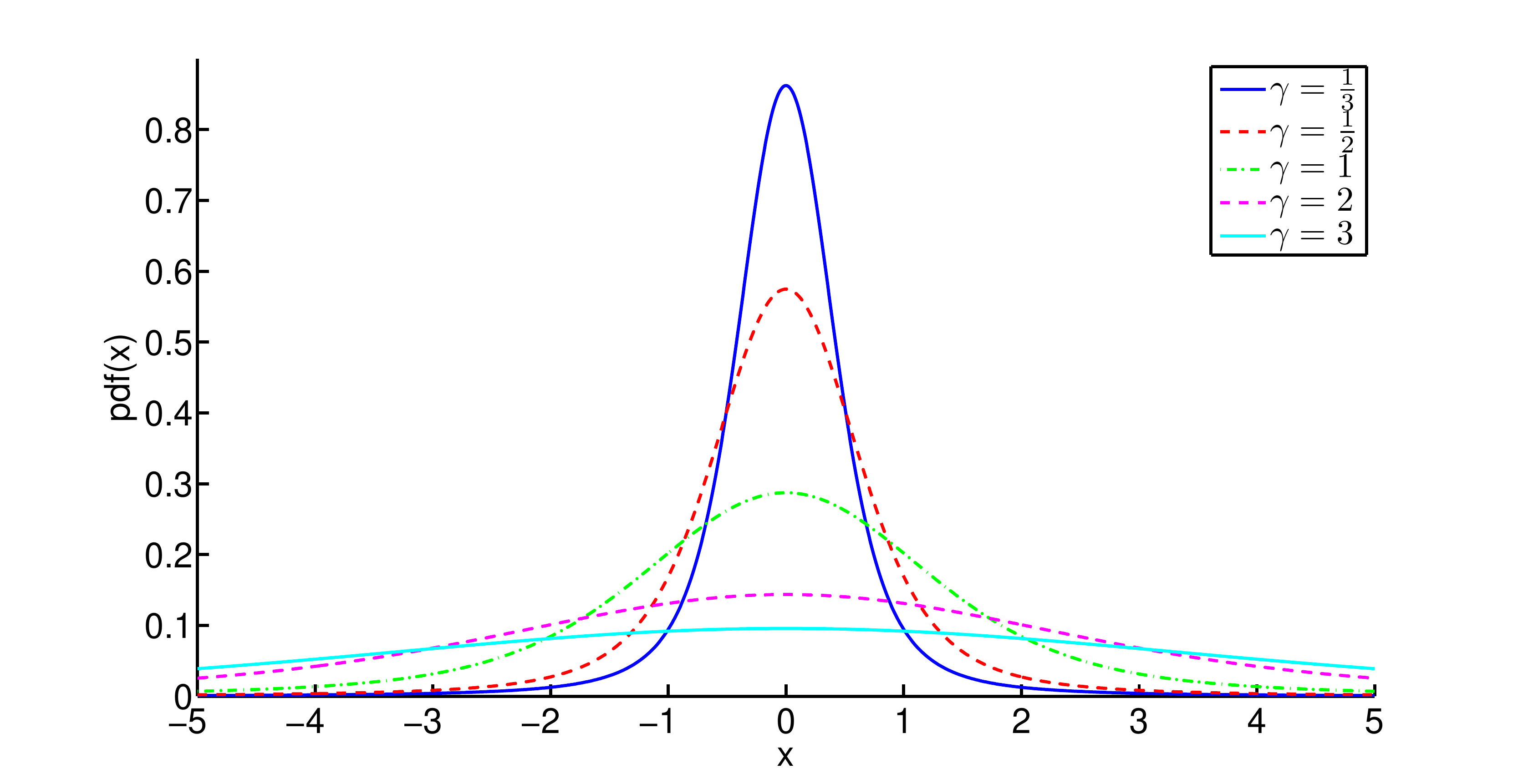}\label{fig:comparaison-gamma}}	&  \subfigure[Influence of parameter $\delta$ with $\alpha=1.5$, $\beta=0$ and $\gamma=1$.]{\includegraphics[height=6cm,width=7cm]{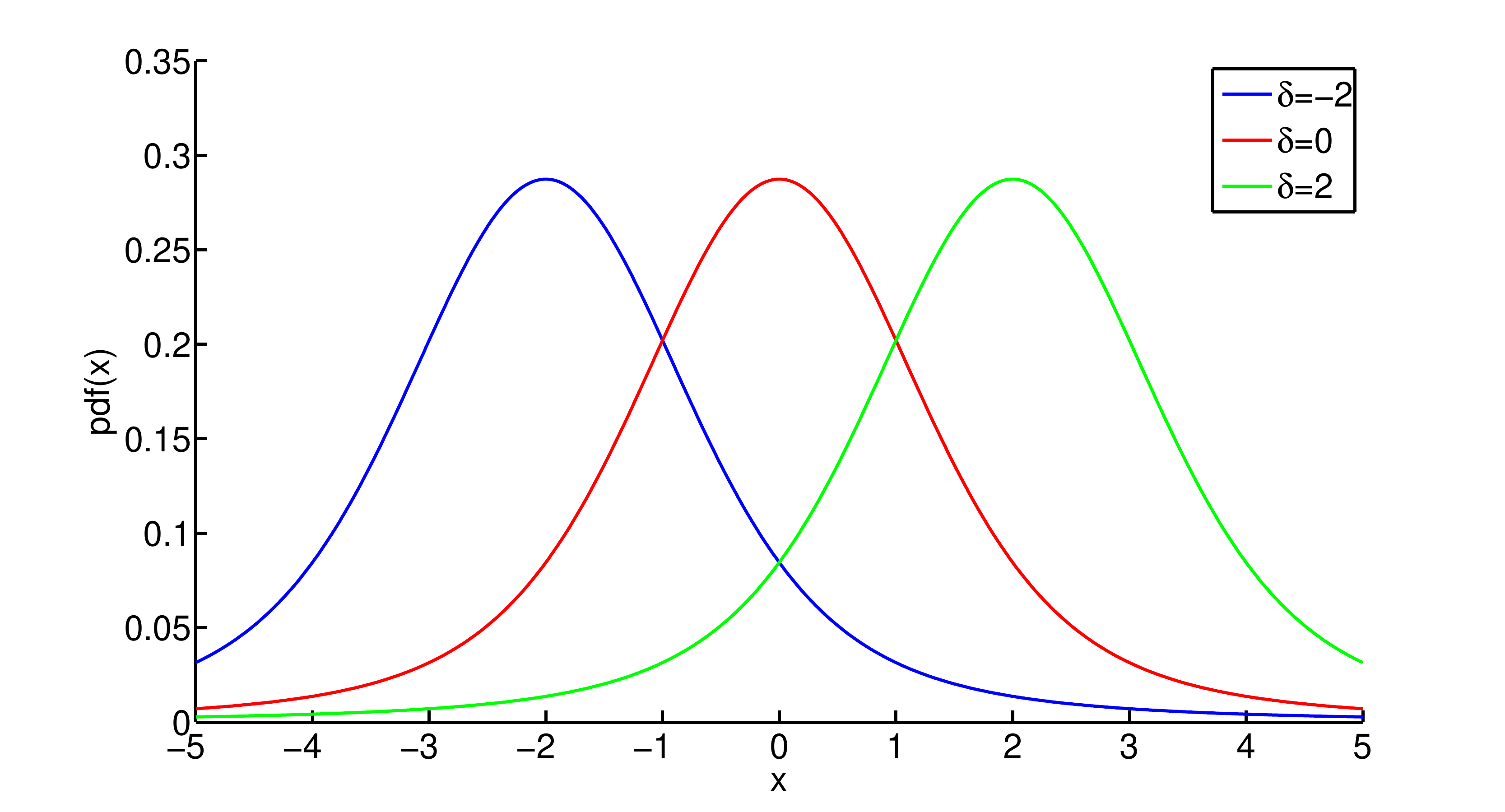}\label{fig:comparaison-delta}}
		\end{tabular}
	\caption{Different $\alpha$-stable {\it pdf}.}
	\label{fig:imagettes}

\end{figure}

However, this definition is problematic for two reasons: while the integrand function is complex, its bounds are infinite. Nolan~\cite{nolan} therefore proposed a way to represent normalized $\alpha$-stable distributions ({\it i.e.} $\gamma=1$ and $\delta=0$). The main idea of Nolan~\cite{nolan} is to use variable modifications so that the integral has finite bounds. Each parameter has an influence on the shape of the $f_{S_{\alpha}(\beta,\gamma,\delta)}$. The curve has a large peak when $\alpha$ is near 0 and a Gaussian shape when $\alpha = 2$ (Figure~\ref{fig:comparaison-alpha}). The shape of the distribution skews to the left if $\beta=1$, to the right if $\beta=-1$ (Figure~\ref{fig:comparaison-beta}) while the distribution is symmetrical when $\beta=0$. Finally, the scale parameter enlarges or compresses the shape of the distribution (Figure~\ref{fig:comparaison-gamma}) and the location parameter leads to the translation of the mode of the $f_{S_{\alpha}(\beta,\gamma,\delta)}$ (Figure~\ref{fig:comparaison-delta}).

\subsection{An overview of $\alpha$-stable estimators}
The estimators of $\alpha$-stable distributions are decomposed into three families: 
\begin{itemize}
\item the sample quantile methods~\cite{fama1971parameter,culloch1980}
\item the sample characteristic function methods~\cite{press1972applied,press1972estimation,koutrouvelis1980regression,koutrouvelis1981iterative}
\item the Maximum Likelihood Estimation~\cite{brorsen1990maximum,mcculloch1998linear,dumouchel1971,nolan2001maximum}.
\end{itemize}

Fama and Roll~\cite{fama1971parameter} developed a method based on quantiles. However, the algorithm proposed by Fama and Roll suffers from a small asymptotic bias in $\alpha$ and $\gamma$ and restrictions on $\alpha \in ]0.6,2]$ and $\beta=0$. McCulloch~\cite{culloch1980} extended the quantile method to the asymmetric case ({\it i.e.} $\beta=0$). The McCulloch estimator is valid for $\alpha \in ]0.6,2]$. 

Press~\cite{press1972applied,press1972estimation} proposed a method based on transformations of characteristic function. In~\cite{weron1995performance}, the author compared the performances of several estimators. For example, the Press method is efficient for specific values. Koutrouvelis~\cite{koutrouvelis1980regression} extended the Press method and proposed a regression method to estimate the parameters of $\alpha$-stable distributions. He proposed a second version of his algorithm which is distinct that it is iterative~\cite{koutrouvelis1981iterative}. In~\cite{akgiray1989estimation}, the authors proved that the method proposed by Koutrouvelis is better than both the quantile method and the Press method because it gives consistent and asymptotically unbiased estimates.   

The Maximum Likelihood Estimation (MLE) was first studied in the symmetric case~\cite{brorsen1990maximum,mcculloch1998linear}. Dumouchel~\cite{dumouchel1971} developed an approximate maximum likelihood method. The MLE was also developed in the asymmetric case. Nolan~\cite{nolan2001maximum} extended the MLE in general case. The problem with the MLE is the calculation of the $\alpha$-stable probability density function because there is no closed-form expression. Moreover, the computational algorithm is time-consuming. 

Consequently, we estimate a univariate $\alpha$-stable distribution using the Koutrouvelis method~\cite{koutrouvelis1981iterative}.
\subsection{Multivariate stable distributions}
It is possible to extend the $\alpha$-stable distributions to the multivariate case.
A random vector ${\bf X} \in \mathbb{R}$ is stable if for all $a, b \in \mathbb{R}^+$, there are $c \in \mathbb{R}^+$ and ${\bf D} \in \mathbb{R}^d$ such that:

\begin{equation}
\displaystyle a {\bf X}_1+b {\bf X}_2=c {\bf X}+ {\bf D},
\end{equation}
where ${\bf X}_1$ and ${\bf X}_2$ are two independent and identically distributed random vectors which follow the same distribution as ${\bf X}$.

The characteristic function of an multivariate $\alpha$-stable distribution, denoted ${\bf X}\sim S_{\alpha,d}(\sigma,{\boldsymbol\delta})$, has the form:
\begin{equation}
\phi_{S_{\alpha,d}({\boldsymbol\sigma},{\boldsymbol\delta})}({\bf t})=\left \lbrace 
\begin{array}{lc}
\displaystyle \exp\left(- \int_{S^{d}} |<{\bf t},{\bf s}>|^\alpha (1-j sgn(<{\bf t},{\bf s}>) \tan({\frac{\pi \alpha}{2}}){\boldsymbol\sigma}(ds) + j <{\boldsymbol\delta},{\bf t}>\right) &\text{ if } \alpha \neq 1,\\
\displaystyle \exp\left(-\int_{S^{d}} |<{\bf t},{\bf s}>| (1+j \frac{\pi}{2}sgn(<{\bf t},{\bf s}>) \ln(<{\bf t},{\bf s}>)){\boldsymbol\sigma}(ds) + j <{\boldsymbol\delta},{\bf t}>\right) &\text{ if } \alpha=1,
\end{array}  
\right.
\end{equation}
with
\begin{itemize}
\item $S^{d}=\{x \in \mathbb{R}^d| ||x||=1$\} the $d$-dimensional unit sphere.
\item ${\boldsymbol \sigma}(.)$ a finite Borel measure on $S^d$. 
\item ${\boldsymbol \delta},{\mathbf t} \in \mathbb{R}^d$.
\end{itemize}
The expression for the characteristic function involves an integration over the unit sphere $S^d$. The measure $\boldsymbol \sigma$ is called the spectral measure and ${\boldsymbol \delta}$ is called the location parameter. The problem with a multivariate $\alpha$-stable distribution is that characteristic function forms a non-parametric set. To avoid this problem, it is possible to consider a discrete spectral measure~\cite{modarres1994method} which has the form:
\begin{equation}
\displaystyle {\boldsymbol\sigma}(.)=\sum_{i=1}^K \gamma_i \delta_{s_i}(.),
\end{equation}
with $\gamma_i$ corresponding to weight and $\delta_{s_i}$ the Dirac measure in $s_i$. For an $\alpha$-stable distribution in $\mathbb{R}^2$ with $K$ mass points, the quantity $\displaystyle s_i=\{\cos(\theta_i),\sin(\theta_i)\}$ with $\theta_i=\frac{2 \pi (i-1)}{K}$. 

There are two methods to estimate an $\alpha$-stable random vector:
\begin{itemize}
\item the PROJection method~\cite{mcculloch2000estimation} (PROJ)
\item the Empirical Characteristic Function method~\cite{nolan2001estimation} (ECF)
\end{itemize}
These two algorithms have the same performances in terms of estimation and computation time.

At this point, it is important to underline that the features extracted from mono-beam echo-sounders have the properties of heavy-tails and skewness. Consequently, we decided to estimate the data with an $\alpha$-stable model. These data are however imprecise and uncertain: the imprecision and uncertainty of data can be linked to poor quality estimation of these data. To take these constraints into consideration, we used an uncertain theory called the theory of belief functions. Consequently, the goal of the next section is to present the theory of belief functions.

\section{The belief functions}
\label{section 3}
The final objective of this paper is to classify synthetic and real data using the theory of belief functions. Data obtained from sensors are generally imprecise and uncertain as noises can disrupt their acquisition. It is possible to consider these constraints by using the theory of belief functions. We will first develop the theory of belief functions within a discrete framework before characterizing it in real numbers.  

\subsection{Discrete belief functions}
This section outlines basic tools in relation to the theory of belief functions.
\subsubsection{Definitions}
Discrete belief functions were introduced by Dempster~\cite{Dempster1967}, and formalized by Shafer~\cite{Shafer1976} where he considers a discrete set of $n$ exclusive events $C_i$ called the frame of discernment:
\begin{equation}
\Theta=\{C_1,\dots,C_n\}.
\end{equation}
$\Theta$ can be interpreted as all the assumptions to a problem. Belief functions are defined as $2^\Theta$ onto [0,1]. The objective of discrete belief functions is to attribute a weight of belief to each element $A\in2^\Theta$. Belief functions must follow the normalization:
\begin{equation}
\sum_{A\subseteq\Theta}m^{\Theta}(A)=1,
\end{equation} 
where $m^{\Theta}$ is called the {\it basic belief assignment} (bba).\\
A focal element is a subset of $A$ where $m^{\Theta}(A)>0$ and several functions  in one-to-one correspondence are built from bba:
\begin{eqnarray}
\hspace{-0.5cm}\displaystyle bel^{\Theta}(A)&=&\sum_{B\subseteq A,B\neq \emptyset}m^{\Theta}(B),\\
\hspace{-0.5cm}\displaystyle pl^{\Theta}(A)&=&\sum_{A\cap B\neq \emptyset}m^{\Theta}(B),\\
\hspace{-0.5cm}\displaystyle q^{\Theta}(A)&=&\sum_{B\subset\Theta,B\supseteq A}m^{\Theta}(B)\label{equation9}.
\end{eqnarray}

The credibility function of $A$ called $bel^{\Theta}(A)$ is all the elements $B\subseteq A$ which believe partially in $A$. This function can be interpreted as a minimum of belief in $A$. On the contrary, a plausibility function $pl^{\Theta}(A)$ illustrates the maximum belief in $A$ while commonality function $q^{\Theta}(A)$ represents the sum of bba allocated to the superset of $A$. This function is very useful as shown below.

\subsubsection{Combination rule}
\label{Combination rule}
We consider $M$ different experts who give mass $m^{\Theta}_i (i=1,\dots,M)$ on each element $A\subseteq\Theta$. Alternatively, it is possible to combine them using combination rules. There are several combination rules~\cite{sentz2002combination} in the literature which differently address conflicts between sources. The most common rule is the conjunctive combination~\cite{smets1990combination} where the resultant mass of $A$ is obtained by:
\begin{equation}
m^{\Theta}(A)=\sum_ {{B_{1}\cap\dots B_{n}=A\neq \emptyset}}\prod_{i=1}^{M} m^{\Theta}_{i}(B_{i})\hspace{0.5cm}, \forall A\in 2^{\Theta}.
\end{equation}
The mass of the empty set is given by:
\begin{equation}
m^{\Theta}(\emptyset)=\sum_ {{B_{1}\cap\dots B_{n}=\emptyset}}\prod_{i=1}^{M} m^{\Theta}_{i}(B_{i})\hspace{0.5cm}, \forall A\in 2^{\Theta}.
\end{equation}
This rule allows us to stay in the open world. However, this is not practical as calculations are difficult. It is possible to calculate resultant mass with communality functions where each mass $m^{\Theta}_i (i=1,\dots,M)$ must be converted into its communality function to calculate the resultant communality function as follows:
\begin{equation}
q^{\Theta}(A)=\prod_{i=1}^{M}q^{\Theta}_{i}(A).
\end{equation}
The final mass is obtained by carrying out the inverse operation of Equation~\eqref{equation9} ~\cite{Shafer1976}.
\subsubsection{Pignistic probability}
To make a decision on $\Theta$, several operators exist such that maximum credibility or maximum plausibility with pignistic probability being the most commonly used operator~\cite{Smets1990}. This name comes from {\it pignus}, a bet, in Latin. This operator approaches the pair ($bel$,$pl$) by uniformly sharing a mass of focal elements on each singleton $C_i$. This operator is defined by:
\begin{equation}
betP(C_{i})=\sum_{A\subset\Theta, C_{i} \in A} \frac{m^{\Theta}(A)}{|A|(1-m^{\Theta}(\emptyset))},
\end{equation}
where $|A|$ represents the cardinality of $A$.\\
We choose the decision $C_i$ by evaluating $\displaystyle \max_{1\leq k\leq n} betP(C_{k})$.
\subsection{Continuous belief functions}
The basic description of continuous belief functions was accomplished by Shafer~\cite{Shafer1976}, then by Nguyen~\cite{nguyen1978random} and Strat~\cite{strat}. Recently, Smets~\cite{Smets2005} extended the definition of belief functions to the set of reals $\overline{\mathbb{R}}=\mathbb{R}\cup\{-\infty,+\infty\}$ and masses are only attributed to intervals of $\overline{\mathbb{R}}$.
\subsubsection{Definitions}
Let us consider $\mathcal{I}=\{[x,y],(x,y],[x,y),(x,y);x,y \in \overline{\mathbb{R}}\}$ as a set of closed, half-opened and opened intervals of $\overline{\mathbb{R}}$.
Focal elements are closed intervals of $\overline{\mathbb{R}}$. The quantity $m^{\mathcal{I}}(x,y)$ are basic belief densities linked to a specific {\it pdf}. If $x>y$, then $m^\mathcal{I}(x,y)=0$. With these definitions, it is possible to define the same functions as in the discrete case.
The interval $[a,b]$ being a set of $\overline{\mathbb{R}}$ with $a\leq b $, the previous functions can be defined as follows:
\begin{eqnarray}
\hspace{-0.5cm}\displaystyle bel^{\overline{\mathbb{R}}}([a,b])&=&\!\!\int_{x=a}^{x=b}\int_{y=x}^{y=b} m^\mathcal{I}(x,y) d\mathit{y} d\mathit{x},\\
\hspace{-0.5cm}\displaystyle pl^{\overline{\mathbb{R}}}([a,b])&=&\!\!\int_{x=-\infty}^{x=b}\int_{y=max(a,x)}^{y=+\infty} m^\mathcal{I}(x,y) d\mathit{y} d\mathit{x},\\
\hspace{-0.5cm}\displaystyle q^{\overline{\mathbb{R}}}([a,b])&=&\!\!\int_{x=-\infty}^{x=a}\int_{y=b}^{y=+\infty} m^\mathcal{I}(x,y) d\mathit{y} d\mathit{x}.
\end{eqnarray}

\subsubsection{Pignistic probability}
The definition of pignistic probability for $a<b$ is:
\begin{equation}
\displaystyle Betf([a,b])=\int_{x=-\infty}^{x=+\infty} \int_{y=x}^{y=+\infty} \frac{|[a,b]\cap[x,y]|}{|[x,y]|} m^\mathcal{I}(x,y) d\mathit{x}d\mathit{y}.
\label{Pignistic probability}
\end{equation}
It is possible to calculate pignistic probabilities to have basic belief densities. However, many basic belief densities exist for the same pignistic probability. To resolve this issue, we can use the consonant basic belief density.
A basic belief density is said to be ``consonant'' when focal elements are nested. Focal elements $I_u$ can be labeled as an index $u$ such that $I_u \subseteq I_u^\prime$ with $u^\prime>u$. This definition is used to apply the least commitment principle, which consists in choosing the least informative belief function when a belief function is not totally defined and is only known to belong to a family of functions. The least commitment principle relies on an order relation between belief functions in order to determine if a belief function is more or less committed than another. For example, it is possible to define on order based on the commonality function:
\begin{equation}
(\forall A\subseteq\Theta, q_1^{\Theta}(A)\leq q_2^{\Theta}(A))\Leftrightarrow(m_1^{\Theta}\subseteq_{q}m_2^{\Theta}).
\end{equation}
The mass function $m_2^{\Theta}$ is less committed than $m_1^{\Theta}$ according to the commonality function.

The function $Betf$ can be induced by a set of isopignistic belief functions ${\mathcal B}{\bf iso}(Betf)$. Many papers~\cite{Smets2005,ristic2005target,caron2006} deal with the particular case of continuous belief functions with nested focal elements. For example, Smets~\cite{Smets2005} proved that the least committed basic belief assignment $m^{\overline{\mathbb{R}}}$ for the commonality ordering attributed to an interval $I=[x,y]$ with $y>x$ related to a bell-shaped~\footnote{{\it i.e.} the {\it pdf} is unimodal with a mode $\mu$, continuous and strictly monotonous increasing (decreasing) at left (right) of the mode.} pignistic probability function is determined by~\footnote{$\delta_d$ refers to the Dirac's measure.}:
\begin{equation}
m^{\overline{\mathbb{R}}}([x,y])=\theta(y) \delta_d(x-\zeta(y)),
\end{equation}    
with $x=\zeta(y)$ satisfying $Betf(\zeta(y))=Betf(y)$ and $\theta(y)$:
\begin{equation}
\theta(y)=(\zeta(y)-y)\frac {d Betf(y)}{d y}.
\label{masse}
\end{equation}
The resultant basic belief assignment $m^{\overline{\mathbb{R}}}$ is consonant and belongs to the set ${\mathcal B}{\bf iso}(Betf)$. However, it is difficult to build belief functions in the particular case of multimodal {\it pdf}s because the frame of discernment has connected sets. In~\cite{doré2009constructing,vannobel2010continuous}, the authors propose a way to build belief functions with connected sets by using a credal measure and an index function. 

\subsection{Credal measure and index function}
\label{Credal measure and index function}
In~\cite{doré2009constructing}, the authors propose a way to calculate belief functions from any probability density function.  
They use an index function $f$ and a specific index space ${\it I}$ to scan the set of focal elements ${\mathcal F}$:
\begin{eqnarray}
f^I:I&\rightarrow& {\mathcal F},\\
y&\longmapsto& f^I(y).
\end{eqnarray}
The authors introduce a positive measure $\mu^\Omega$ such that $\int_I d\mu^{\Omega}(y)\leq1$ describes unconnected sets. The pair $(f^I,\mu^\Omega)$ defines a belief function. For all $A\in \Omega$, they define subsets which belong to the Borel set:
\begin{eqnarray}
F_{\subseteq A}&=&\{y\in I|f^I(y)\subseteq A\},\\
F_{\cap A}&=&\{y\in I|f^I(y)\cap A\neq\emptyset\},\\
F_{\supseteq A}&=&\{y\in I|f^I(y)\supseteq A\}.
\end{eqnarray}  
From these definitions, they compute belief functions:
\begin{eqnarray}
\displaystyle bel^{\Omega}(A)&=&\int_{F_{\subseteq A}} d\mu^{\Omega}(y),\\
\displaystyle pl^{\Omega}(A)&=&\int_{F_{\cap A}} d\mu^{\Omega}(y),\\
\displaystyle q^{\Omega}(A)&=&\int_{F_{\supseteq A}} d\mu^{\Omega}(y).
\end{eqnarray}

They continue their study by considering consonant belief functions. The set of focal elements ${\mathcal F}$ must be ordered from the operator $\subseteq$. They define an index function $f$ from $\mathbb{R}^+$ to ${\mathcal F}$ such that:
\begin{equation}
y\geq x \Longrightarrow f(y) \subseteq f(x) . 
\end{equation}
They generate consonant sets by using a continuous function $g$ from $\mathbb{R}^d$ to $I=[0,\alpha_{max}[$. The $\alpha$-cuts are the set:
\begin{equation}
f_{cs}^I =\{x \in \mathbb{R}^d|g(x)\geq \alpha \}.
\end{equation}
They finally define the index function:
\begin{eqnarray}
f_{cs}^I: I=[0,\alpha_{max}]&\rightarrow&\{ f_{cs}^I(\alpha)|\alpha \in I\},\\
\alpha&\longmapsto& f_{cs}^I(\alpha).
\end{eqnarray}
The information available is the conditional pignistic density $Betf$. However, many basic belief densities exist for the same pignistic probability $Betf$. The least commitment principle allows the least informative basic belief density to be chosen, where the focal elements are the $\alpha$-cuts of $Betf$ such that~\footnote{$\lambda$ refers to the Lebesgue's measure.}:
\begin{equation}
d\mu^{\Omega}(y)(\alpha)=\lambda(f_{cs}^I(\alpha)) d\lambda(\alpha).
\end{equation}

\section{Continuous belief functions and $\alpha$-stable distribution}
\label{section 4}
In this section, we model data distributions as a single $\alpha$-stable distribution. We must however introduce the notion of plausibility function for an $\alpha$-stable distribution. We first describe how to calculate the plausibility function knowing the {\it pdf} in $\overline{\mathbb{R}}$ before we extend to in $\overline{\mathbb{R}}^d$. We finally explain how we construct our belief classifier.

\subsection{Link between pignistic probability function and plausibility function in $\overline{\mathbb{R}}$}

The information available is the conditional pignistic density $Betf[C_i]$ with $C_i \in \theta$. The function $Betf[C_i]$ is supposed to be bell-shaped for all $\alpha$-stable distributions (proved by Yamazato~\cite{yamazato1978unimodality}). 

The plausibility function from a mass $m^{\overline{\mathbb{R}}}$ is obtained by an integral of Equation~\eqref{masse} between $[x,+\infty[$:
\begin{equation}
\displaystyle pl^{\overline{\mathbb{R}}}[C_i](I)=\int_{x}^{+\infty}(\zeta(t)-t)\frac{d Betf(t)}{d t} d t.
\label{masse1} 
\end{equation}
By assuming that $Betf$ is symmetrical, an integration by parts can simplify Equation~\eqref{masse1} :
\begin{equation}
\displaystyle pl^{\overline{\mathbb{R}}}[C_i](I)=2(x-\mu) Betf(x) + 2 \int_{x}^{+\infty}Betf(t) d t. 
\label{pl}
\end{equation}
Now let us consider a particular case where symmetrical $Betf$ is an $\alpha$-stable distribution (the parameter $\beta=0$). We already know that $ Betf(x)=f_{S_{\alpha}(\beta,\gamma,\delta)}(x)$. We can calculate $ \int_{x}^{+\infty}Betf(t) d t$ by using the Chasles' theorem:
\begin{equation}
\displaystyle \int_{-\infty}^{+\infty}f_{S_{\alpha}(\beta,\gamma,\delta)}(t) d t=\int_{-\infty}^{x}f_{S_{\alpha}(\beta,\gamma,\delta)}(t) d t +\int_{x}^{+\infty}f_{S_{\alpha}(\beta,\gamma,\delta)}(t) d t.
\end{equation}
By definition, a $pdf$ has the quantity $\int_{-\infty}^{+\infty}f_{S_{\alpha}(\beta,\gamma,\delta)}(t) d t=1$ and $\int_{-\infty}^{x}f_{S_{\alpha}(\beta,\gamma,\delta)}(t) d t$ represents the definition of the $\alpha$-stable cumulative density function $F_{S_{\alpha}(\beta,\gamma,\delta)}$.

Consequently, Equation~\eqref{pl} can be simplified:
\begin{equation}
pl^{\overline{\mathbb{R}}}[C_i](I)=2(x-\mu) f_{S_{\alpha}(\beta,\gamma,\delta)}(x)+2(1-F_{S_{\alpha}(\beta,\gamma,\delta)}(x)).
\end{equation}
The plausibility function related to an interval $I=[x,y]$ can also be seen as the area defined under the $\alpha_{cut}$-cut such that $\alpha_{cut}=Betf(x)$. The notation $pl[C_i](x)$ is equivalent to $pl[C_i](I)$.

\begin{figure}[ht]
	\centering
		\includegraphics[scale=0.25]{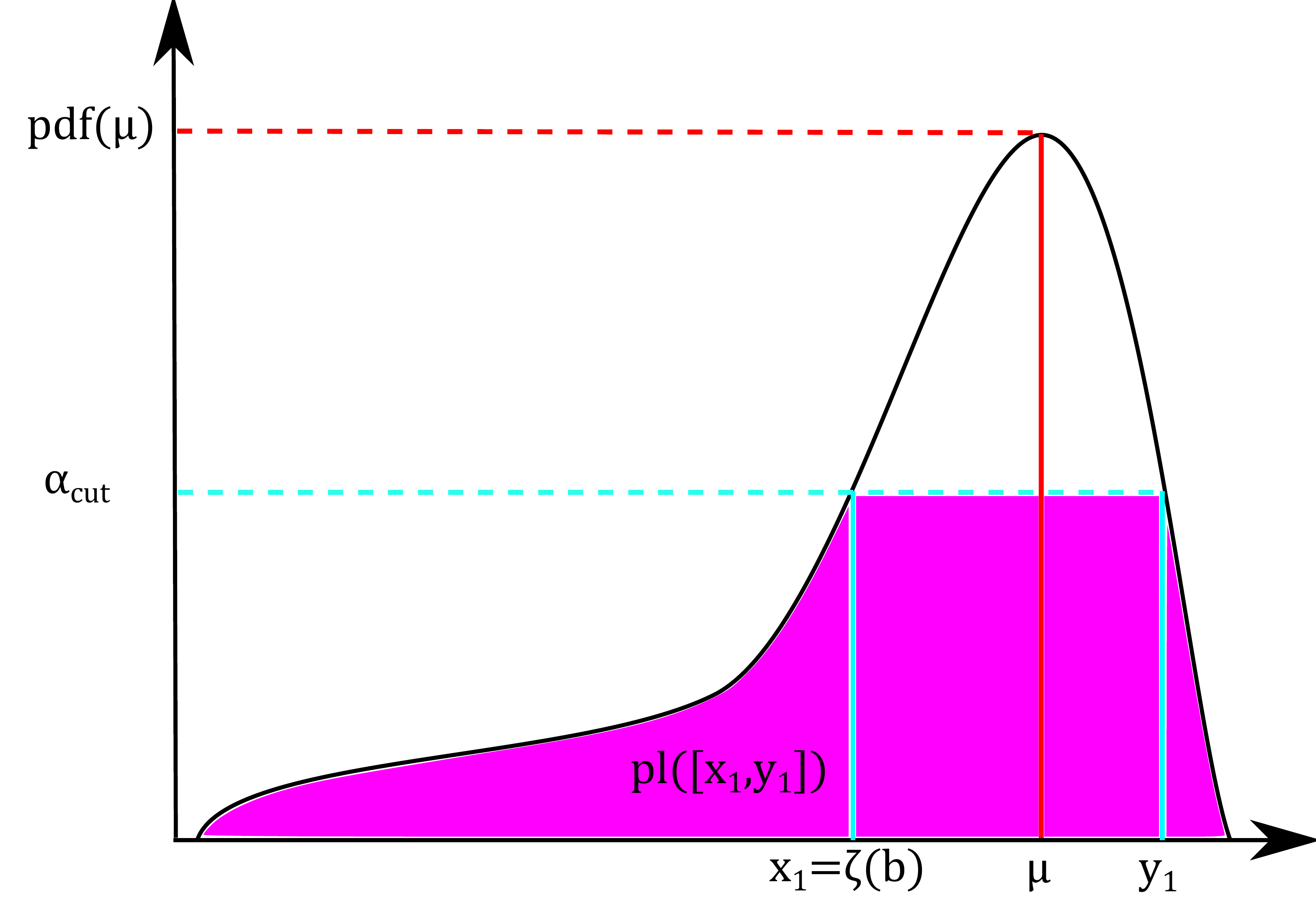}
	\caption{Calculation of plausibility function with an unsymmetric $\alpha$-stable distribution.}
	\label{fig:dessin_pl_non_sym}
\end{figure}

Now us let consider an asymmetric $\alpha$-stable probability density function. We must proceed numerically to calculate plausibility function at point $x_1>\mu$, with $\mu$ the mode of the probability density function (Figure~\ref{fig:dessin_pl_non_sym}). The plausibility function related to an interval $I_1=[x_1,y_1]$ is defined by the area defined under the $\alpha_{cut}$-cut such that $\alpha_{cut}=Betf(x_1)$:
\begin{equation}
\displaystyle pl^{\overline{\mathbb{R}}}[C_i](I_1)=\int_{-\infty}^{x_1}Betf(t) dt +(y_1-x_1)Betf(x_1)+\int_{y_1}^{+\infty} Betf(t) dt.
\end{equation}
By definition, a $pdf$ has the quantity $ \int_{y_1}^{+\infty}f_{S_{\alpha}(\beta,\gamma,\delta)}(t) d t=1-F_{S_{\alpha}(\beta,\gamma,\delta)}(y_1)$ and $\int_{-\infty}^{x_1}f_{S_{\alpha}(\beta,\gamma,\delta)}(t) d t=F_{S_{\alpha}(\beta,\gamma,\delta)}(x_1)$.
In general, we know only one point $y_1$. We estimate numerically $x_1$ such that $f_{S_{\alpha}(\beta,\gamma,\delta)}(y_1)=f_{S_{\alpha}(\beta,\gamma,\delta)}(x_1)$. Finally, the plausibility function related to the interval $I_1$ is:
\begin{equation}
pl^{\overline{\mathbb{R}}}[C_i](I_1)=1+F_{S_{\alpha}(\beta,\gamma,\delta)}(x_1)-F(y_1)+(y_1-x_1)f(x_1).
\label{eq_plaus_stable}
\end{equation}
In practice, we base the classification on several features defined for different classes $\Theta=\{C_1,..,C_n\}$. For example, the Haralick parameters ``contrast'' and ``homogeneity'' have different values for the classes rock, silt and sand. The features are modeled by a probability density function because the features have continuous values. We can calculate a plausibility function related to its probability density functions by using the least commitment principle. Several plausibility functions associated to the same feature can be combined by using the generalized Bayes theorem~\cite{Smets1993,Delmotte2004} to calculate mass functions allocated to $A$ of an interval $I$:
\begin{equation}
\displaystyle m^{\overline{\mathbb{R}}}[x](A)= \prod_{C_j\in A} pl_{j}(x)\displaystyle { \prod_{C_j\in A^c} (1-pl_{j}(x))} .
\label{theo_bayes}
\end{equation}
For several features, it is possible to combine the mass functions with combination rules.

\begin{figure}[!h]
	\centering
		\includegraphics[scale=0.5]{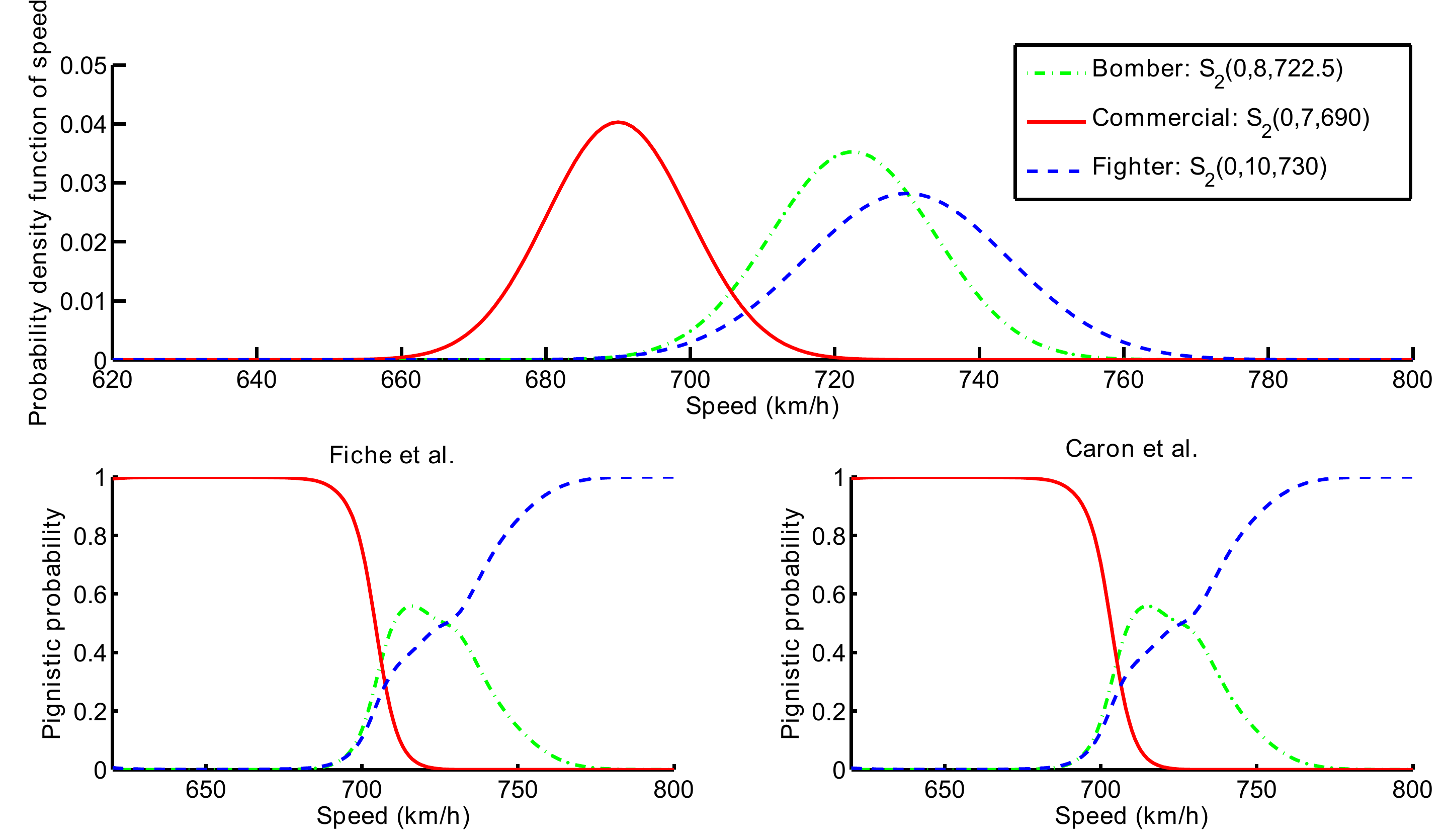}
	\caption{Representation of probability density function of speed and the pignistic probabilities calculated from the Fiche {\it et al.} and Caron {\it et al.}~\cite{caron2006} approaches.}
	\label{fig:fig_avion_eng}
\end{figure}

To validate our approach, we classify planes using kinematic data as in~\cite{ristic2005target,caron2006} and compare the decision with the approach of Caron {\it et al.}~\cite{caron2006}. We associate a Gaussian probability density function for each speed's target (Figure~\ref{fig:fig_avion_eng}):
\begin{itemize}
\item Commercial defined by the probability density function of speed $S_2(0,8,722.5)$.
\item Bomber defined by the probability density function of speed $S_2(0,7,690)$.
\item Fighter defined by the probability density function of speed $S_2(0,10,730)$.
\end{itemize}
We can observe that the decision is the same with the both approaches in the particular case of Gaussian probability density functions.
\subsection{Link between pignistic probability function and plausibility function to $\overline{\mathbb{R}}^d$}
  
\begin{figure}[h]
	\centering
		\includegraphics[scale=0.25]{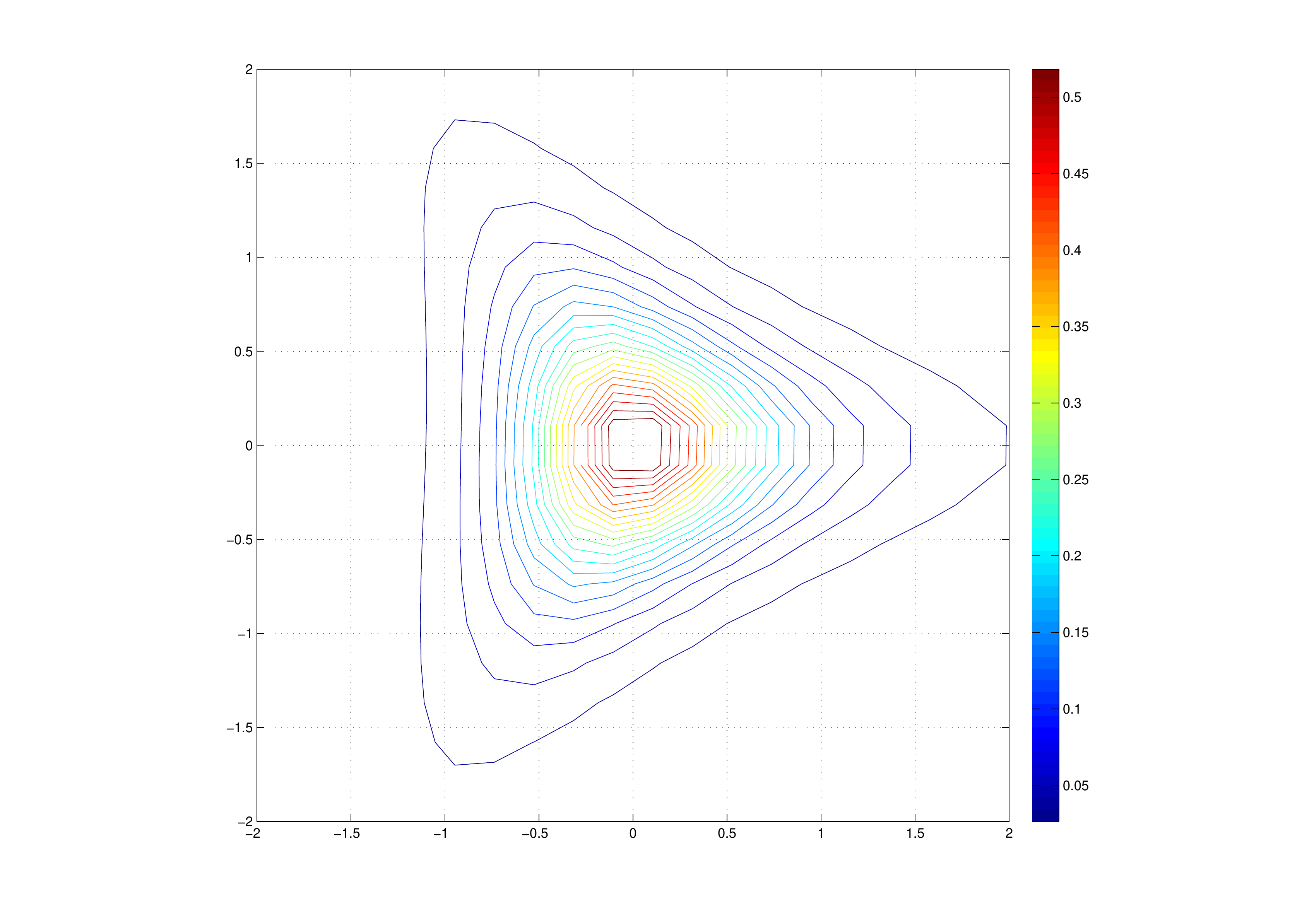}
	\caption{Isoprobability surface of $f$.}
	\label{fig:example_methode_ecf}
\end{figure}

It is possible to extend plausibility function in $\overline{\mathbb{R}}^d$. In~\cite{caron2006}, the authors calculate plausibility function in the Gaussian {\it pdf} situation of mode ${\boldsymbol \delta}$ and matrix of covariance ${\boldsymbol \Sigma}$. Mass function is built in such a way that isoprobability surfaces $S_i$ with $1\leq i\leq n$ are focal elements. In $\mathbb{R}^2$, isoprobability points of a multivariate Gaussian pdf of mean ${\boldsymbol \delta}$ and covariance matrix ${\boldsymbol \Sigma}$ are ellipses. In dimension $d$, the authors defined focal elements as the nested sets $HV_\alpha$ enclosed by the isoprobability hyperconics $HC_\alpha=\{{\bf x}\in \mathbb{R}^d | ({\bf x}-{\boldsymbol \delta)}\,^t {\boldsymbol \Sigma}^{-1}({\bf x}-{\boldsymbol \delta)}\}$. They obtain bbd by applying the least commitment principle:
\begin{equation}
\displaystyle m^{\overline{\mathbb{R}}^d}(HV_\alpha)=\frac{\alpha^{\frac{d+2}{2}-1}}{2^{\frac{d+2}{2}} \Gamma(\frac{d+2}{2})} \exp{(-\frac{1}{2}\alpha}) \text{ with $\alpha \geq0$}.
\label{equation chi2}
\end{equation}
Equation~\eqref{equation chi2} defines a $\chi^2$ distribution with $d+2$ degrees of freedom. The plausibility function at point $\bf{x}$ belonging to a surface $S_i$ corresponds to the volume delimited by $S_i$ (we give an example in Figure~\ref{fig:example_methode_ecf} for an $\alpha$-stable {\it pdf}). Consequently, the plausibility function at point ${\bf x}$ is defined by:
\begin{equation}
\displaystyle pl^{\overline{\mathbb{R}}^d}({\bf x}\in \mathbb{R}^{d})=\int_{\alpha=({\bf x}-{\boldsymbol \delta})\,^t {\boldsymbol \Sigma}^{-1}({\bf x}-{\boldsymbol \delta})}^{\alpha=+\infty} \frac{\alpha^{\frac{d+2}{2}-1}}{2^{\frac{d+2}{2}} \Gamma(\frac{d+2}{2})} \exp{(-\frac{1}{2}\alpha}) d\alpha.
\label{equation chi2 bis} 
\end{equation}
Equation~\eqref{equation chi2 bis} can be simplified by:
\begin{equation}
\displaystyle pl^{\overline{\mathbb{R}}^d}({\bf x}\in \mathbb{R}^{d})=1-F_{d+2}(({\bf x}-{\boldsymbol \delta})({\boldsymbol \Sigma})^{-1}({\bf x}-{\boldsymbol \delta})).
\label{equation_Gauss}
\end{equation}
The function $F_{d+2}$ is a cumulative density function of the $\chi^2$ distribution with $d+2$ degrees of liberty ($d$ is the dimension of vector {\bf x}).

The authors also calculate plausibility functions in the particular case of GMMs with their {\it pdf}s denoted by $p_{GMM}$:
\begin{equation}
\displaystyle p_{GMM}({\bf x})=\sum_{k=1}^{k=n} w_k {\cal N}({\bf x},{\boldsymbol \delta}_k,{\boldsymbol \Sigma}_k),
\end{equation}
where ${\cal N}(x,{\boldsymbol \delta}_k,{\boldsymbol \Sigma}_k)$ is the normal distribution of the $k$ components with mean ${\boldsymbol \delta}_k$ and matrix of covariance ${\boldsymbol \Sigma}_k$ and $w_k$ is the weight of each mixture.
They assign belief to nested sets belonging to a component. Consequently, the plausibility function of the GMMs can be seen as a weighted sum of plausibility functions defined by each component of the mixture:
\begin{equation}
\displaystyle pl^{\overline{\mathbb{R}}^d}({\bf x}\in \mathbb{R}^{d})=1-\sum_{k=1}^{k=n} w_k F_{d+2}(({\bf x}-{\boldsymbol \delta}_k)({\boldsymbol \Sigma}_k)^{-1}({\bf x}-{\boldsymbol \delta}_k)).
\end{equation}

We want to extend the calculation of plausibility function for any $\alpha$-stable {\it pdf}. However, there is no closed-form expression for $\alpha$-stable {\it pdf}. We use the approach developed by Dor\'e {\it et al.}~\cite{doré2009constructing} to build belief functions.

\subsection{Belief classifier}
Let us consider a data set with $N$ samples from $d$ sensors. For example, each feature of a vector ${\bf x}~\in~\mathbb{R}^d$ can be seen as a piece of information from a sensor. The classification is divided into two steps. $N \times p$ (with $p \in ]0,1[$) samples are first picked out randomly from the data set for the learning base, noted ${\bf X}$, such that:
\begin{equation*}
{\bf X}=
\begin{pmatrix}
x_1^1&\dots&x_1^{N\times p}\\
\vdots&\vdots&\vdots\\
x_d^1&\dots&x_d^{N\times p}\\
\end{pmatrix}
\end{equation*}

\begin{figure}[h]
	\centering
		\includegraphics[scale=0.5]{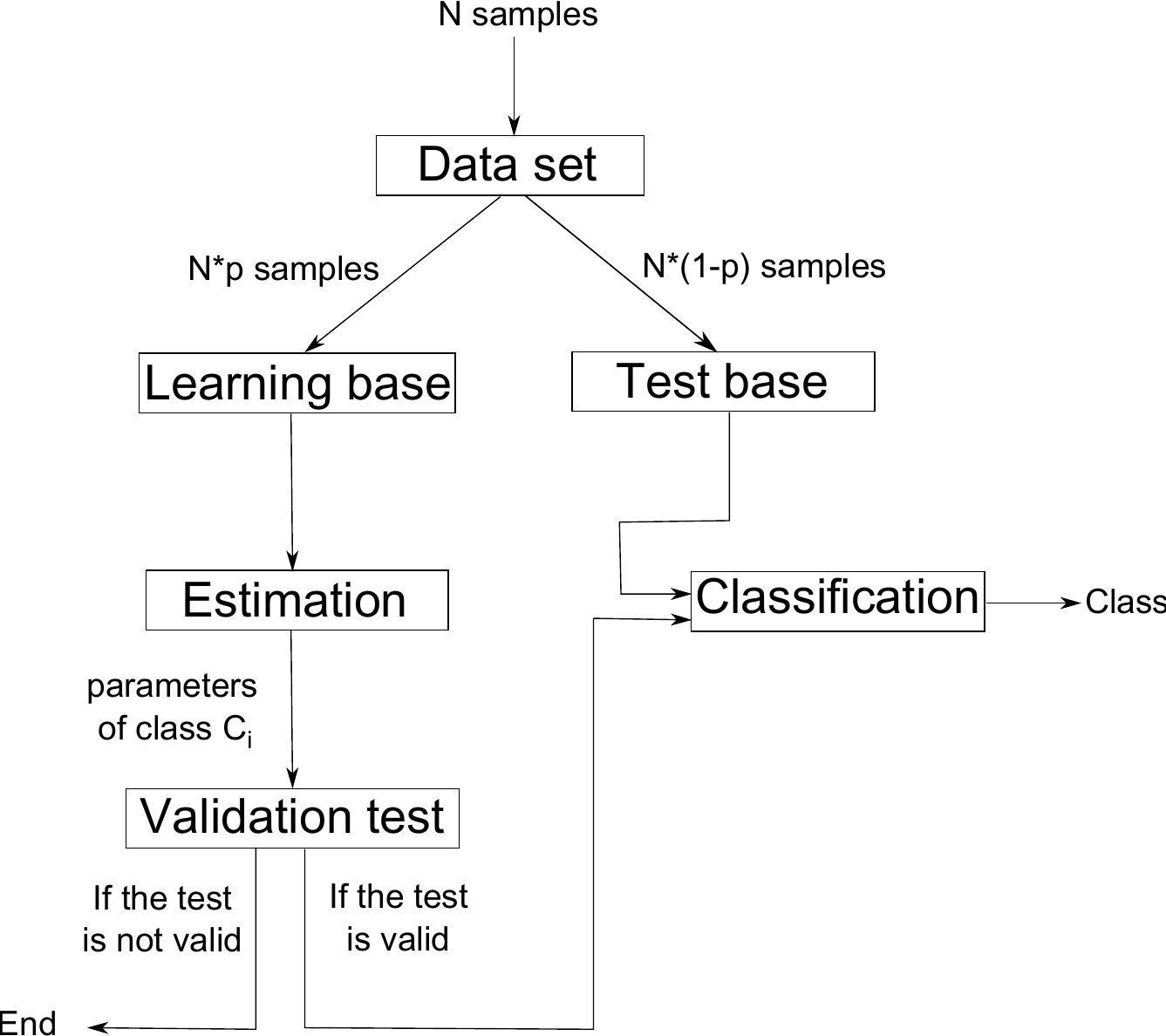}
	\caption{Scheme of classification.}
	\label{fig:classif_schema}
\end{figure}

\begin{figure}[ht]
	\centering
		\includegraphics[height=10cm,width=15cm]{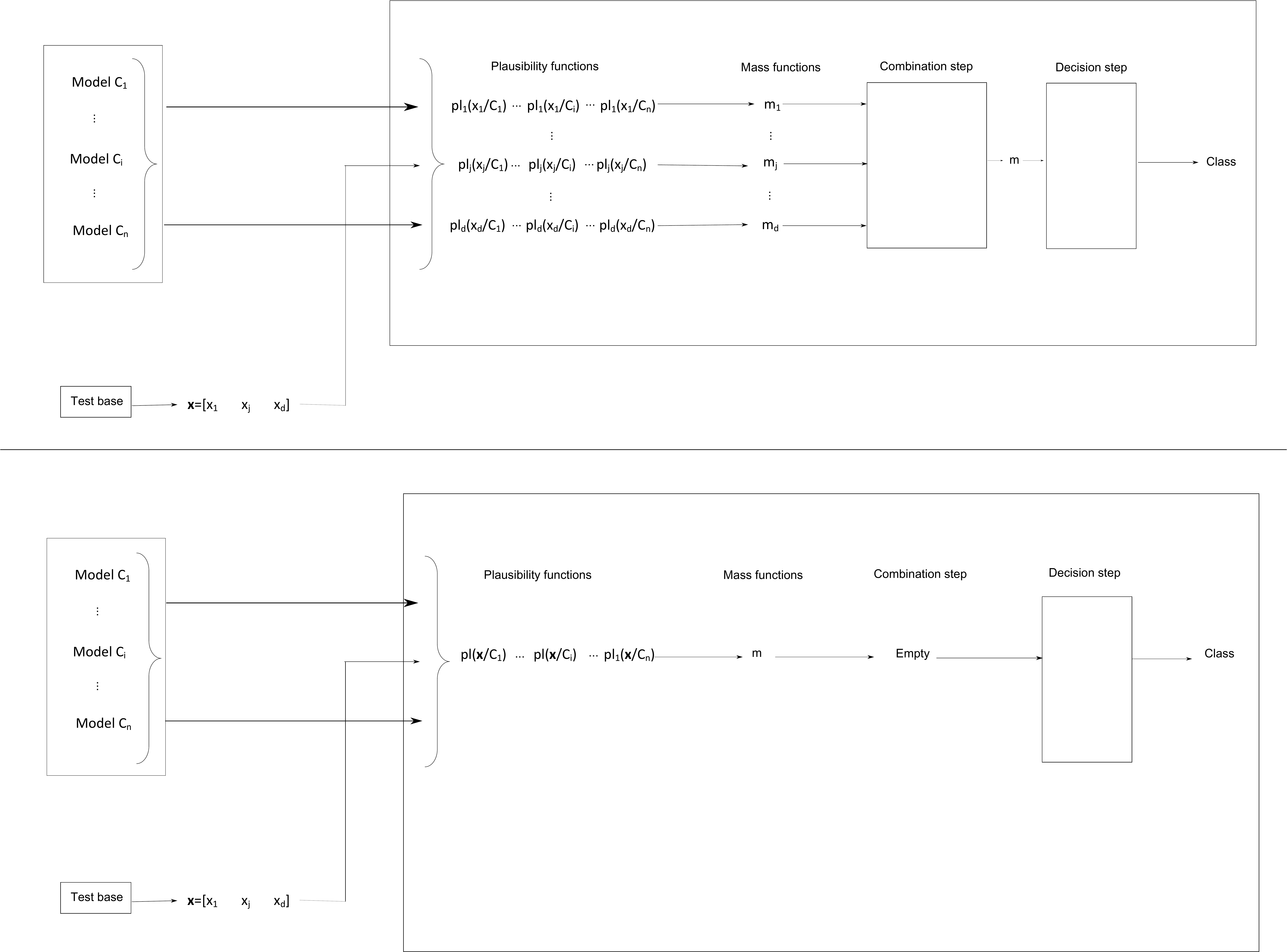}
	\caption{Scheme of classification step in one and $d$ dimensions.}
	\label{fig:schemabis}
\end{figure}

All columns of ${\bf X}$ belong to a class $C_i \text{ with } 1\leq i \leq n$. Probability density functions (Gaussian or $\alpha$-stable models) are estimated from samples belonging to classes. The rest of the samples is used for the test base ($N\times (1-p)$ vectors). We use a validation test to determine if samples belong to an estimated model. If the test is not valid, we stop the classification, otherwise we continue the classification (Figure~\ref{fig:classif_schema}). The classification step diagram in one and $d$ dimensions is shown in Figure~\ref{fig:schemabis}.

Plausibility functions knowing classes for $x$ are calculated either from their probability density functions by using Equation~\eqref{eq_plaus_stable} for an unsymmetric $\alpha$-stable probability density function or by using Equation~\eqref{equation_Gauss} for a Gaussian probability density function. We obtain $d$ mass functions (one for each feature) at point $x$ with the generalized Bayes theorem~\cite{Smets1993,Delmotte2004} (Equation~\eqref{theo_bayes}). These $d$ mass functions are combined by a combination rule to obtain a single mass function (section~\ref{Combination rule}). Finally, the mass function is transformed into pignistic probability (Equation~\eqref{Pignistic probability}) to make the decision.

It is also possible to work with a vector $\mathbf x$ of $d$ dimensions. For an $\alpha$-stable probability density function, plausibility functions are calculated for each feature by using the approach of Dor{\'e} {\it et al}~\cite{doré2009constructing} (section~\ref{Credal measure and index function}). For a Gaussian probability density function, we use Equation~\eqref{equation_Gauss} to calculate plausibilty functions. We calculate one mass function at point $\mathbf x$ by using the generalized Bayes theorem (Equation~\eqref{theo_bayes}). There is no combination step because we are calculating a single mass function. We use Equation~\eqref{Pignistic probability} to transform the mass function into pignistic probabilities. The decision is chosen by using the maximum number of pignistic probabilities.

\section{Application to pattern recognition}
\label{section 5}
The aim is to build a belief classifier and to perform a classification of synthetic and real data by estimating features using a Gaussian and $\alpha$-stable {\it pdf}. In~\cite{fiche2010}, synthetic data are classified by modeling features using Gaussian and $\alpha$-stable mixture models. By observing confidence intervals, the hypothesis of $\alpha$-stable mixture models is significantly better than the hypothesis of Gaussian mixture models. However, when the number of Gaussian distributions increases, classification accuracies are significantly the same. Images from a side-scan sonar are automatically classified by extracting Haralick features~\cite{haralick1973textural,haralick2005statistical}. However, classification accuracies are roughly the same.

In this section, we limit algorithms with a vector of features in dimension $d\leq2$. Indeed, the generalization of $\alpha$-stable {\it pdf} in dimension $d>2$ increases CPU time because there is no closed-form expression. Consequently, we distinguish two cases during the classification step:
\begin{itemize}
\item The one dimension case: each dimension is considered as a feature.
\item The two  case: we considered a vector of two features.
\end{itemize} 

We also use a statistical test called Kolmogorov-Smirnov test (K-S test) to evaluate the quality of each model. A Kolmogorov-Smirnov test (K-S test) with a significance level of 5~\% is also used to determine if two datasets differ significantly. 
Let us assume two null hypotheses:
\begin{itemize}
\item $H_1$: ``Data follow a Gaussian distribution''.
\item $H_2$: ``Data follow an $\alpha$-stable distribution''.
\end{itemize}
The notation $H_1$ and $H_2$ are used for the rest of the paper. If the test rejects a null hypothesis $H_i$, we stop the classification. Otherwise, if the test is valid, we then classify the samples belonging to the test base by using the belief classifier. 

The results obtained with the belief functions are compared using a Bayesian approach. The prior probability $p(C_i)$ is first calculated corresponding to the proportion of each class in the learning set. For each class, the application of Bayes theorem gives posterior probabilities:
\begin{equation}
\displaystyle p(C_i/x)=\frac{p(x/C_i)p(C_i)}{\displaystyle \sum_{u=1}^{n}p(x/C_u) p(C_u)}.
\end{equation}
Finally, the decision is chosen by using the maximum nimber of posterior probabilities. 

We first present a classification of synthetic data, belonging to Gaussian distributions, by modeling features using a Gaussian distribution and $\alpha$-stable {\it pdf}. We then compare the $\alpha$-stable and Gaussian model with synthetic data belonging to $\alpha$-stable distributions. The same approach is finally applied to real data from a mono-beam echo-sounder. 
\subsection{Application to synthetic data generated by Gaussian distributions}
In this subsection, we show that the $\alpha$-stable and Gaussian model have the same behavior during the classification when synthetic data are generated by Gaussian distributions.
\subsubsection{Presentation of the data}

\begin{figure}[ht]
	\centering
		\includegraphics[scale=0.53]{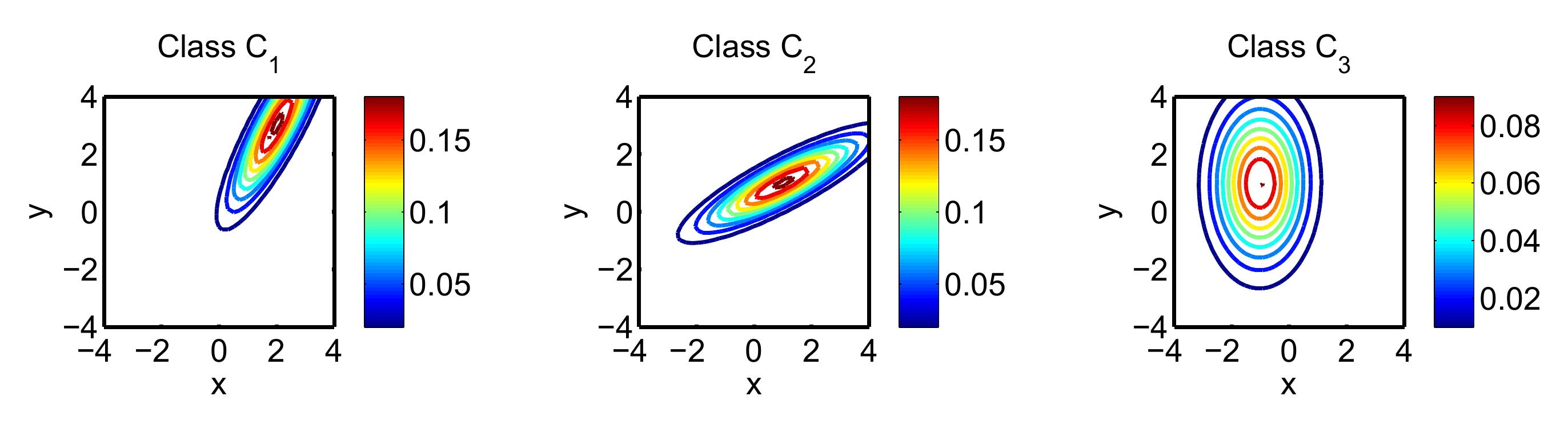}
	\caption{Iso-contour lines of generated Gaussian pdf.} 
	\label{fig:gen_gauss}
\end{figure}

\begin{table*}[h!]    
\begin{center}
\begin{tabular}{|c|c|c|} 
\cline{2-3}
\multicolumn{1}{c|}{} & ${\boldsymbol\mu}$& $V$\\
\hline
First Gaussian {\it pdf} ($C_1$)&$\displaystyle\begin{pmatrix}2 \\3\end{pmatrix}$ & $\displaystyle\begin{pmatrix}1&1.5\\1.5&3\end{pmatrix}$ \\
\hline
Second Gaussian {\it pdf} ($C_2$)& $\begin{pmatrix}1 \\1\end{pmatrix}$ &  $\displaystyle\begin{pmatrix}3&1.5 \\1.5&1\end{pmatrix}$\\
\hline
Third Gaussian {\it pdf} ($C_3$)& $\displaystyle\begin{pmatrix}-1 \\1\end{pmatrix}$& $\begin{pmatrix}1 &0\\0 &3\end{pmatrix}$ \\
\hline
\end{tabular}
\caption{Numerical values for the Gaussian {\it pdf}.}
\label{table_gaussian}
\end{center}
\end{table*}
We simulated three Gaussian {\it pdf}s in $\mathbb{R} \times \mathbb{R}$ ({\it c.f.} Table~\ref{table_gaussian}). Each distribution holds 3000 samples. A mesh between $[-4,4]\times [-4,4]$ enables the level curves of {\it pdf} to be plotted (Figure~\ref{fig:gen_gauss}).

\subsubsection{Results}

We generated data and split samples randomly into two sets: one third for the learning base and the rest for the test base. The learning base was used to estimate each mean and variance for the Gaussian distributions and each parameter $\alpha$, $\beta$, $\gamma$ and $\delta$ for the $\alpha$-stable distributions~\cite{fama1965behavior,koutrouvelis1980regression,mcculloch1986simple}. We needed to make a mesh between $[-4,4]\times [-4,4]$ to calculate plausibility functions in $\mathbb{R}^2$ with the $\alpha$-stable model. Consequently, samples were chosen possessing features with dimensions within $[-4,4]$.   

\paragraph{The one dimension case}
~~\\
For each class $C_i$, we consider each feature as a source of information. The estimated probability density functions can be seen as pignistic probability functions. For each pignistic probability function, we associate a mass function which will subsequently be combined with other mass functions.

\begin{table*}[htbp]    
\begin{center}
\begin{tabular}{|c|c|c|c|c|c|} 
\cline{3-6}
\multicolumn{2}{c|}{} & \multicolumn{2}{|c|}{$H_1$}&\multicolumn{2}{|c|}{$H_2$}\\
\cline{3-6}
\multicolumn{2}{c|}{} & p-value&ksstat&p-value&ksstat\\
\hline
$C_1$& first feature& 0.7021& 0.0542&0.9241&0.0422\\
\cline{2-6}
 & second feature& 0.8799 &0.0452&0.9241&0.0422\\
 \hline
 $C_2$& first feature& 0.8693& 0.0464&0.3736&0.0712\\
\cline{2-6}
 & second feature& 0.6855 &0.0557&0.3736&0.0712\\
 \hline
 $C_3$& first feature& 0.8437& 0.0464&0.4150&0.0667\\
\cline{2-6}
 & second feature& 0.6602 &0.0551&0.7866&0.0493\\
 \hline
\end{tabular}
\caption{Statistical values for a Kolmogorov-Smirnov test with a significance level of 5~\% (p-value: the critical value to reject the null hypothesis, ksstat: the greatest discrepancy between the observed and expected cumulative frequencies).}
\label{KS_test_gaussien}
\end{center}
\end{table*}

\begin{figure}[ht]
	\centering
		\includegraphics[scale=0.4]{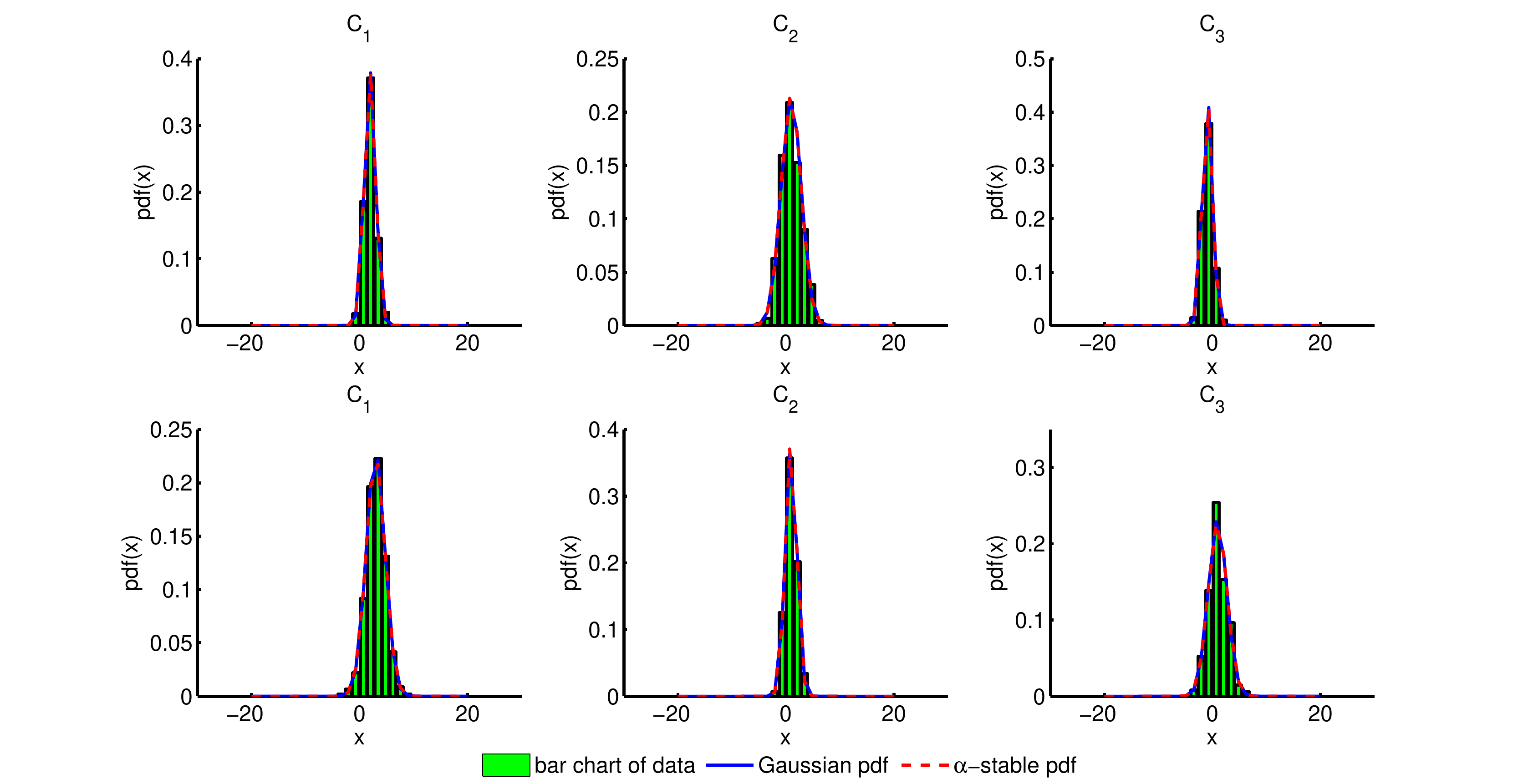}
	\caption{Empirical {\it pdf} and its estimations (The first row corresponds to the first feature and the second row corresponds to the second feature).}
	\label{fig:gaussian_1D_comparaison}
\end{figure}

The K-S test is valid for the two null hypotheses $H_1$ and $H_2$ (Table~\ref{KS_test_gaussien}). We can assert that samples are drawn from the same distribution. The proposed method performs well in the particular case of synthetic data belonging to Gaussian distributions. However, the classification rates (Table~\ref{comparaison_classif}) are very low because there is confusion between classes (Figure~\ref{fig:gaussian_1D_comparaison}). The classification accuracy obtained with the Bayesian approach is significantly the same as the belief functions. Indeed, the Bayesian performs well provided that the probability density functions are well-estimated. The classification results depend on the prior probabilitiy estimation: if a prior probability of a class is underestimated, the confusion between classes increases.  

\begin{figure}[ht]
	\centering
		\includegraphics[scale=0.53]{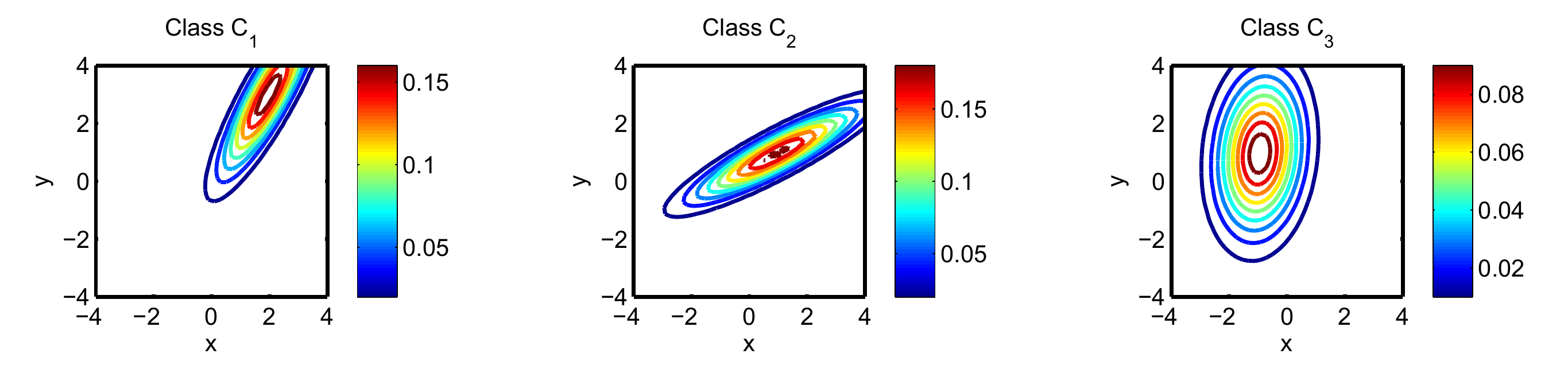}
	\caption{Iso-contour lines of data estimated with a Gaussian model.}
	\label{fig:est_gauss}
\end{figure}

\begin{figure}[ht]
	\centering
		\includegraphics[scale=0.53]{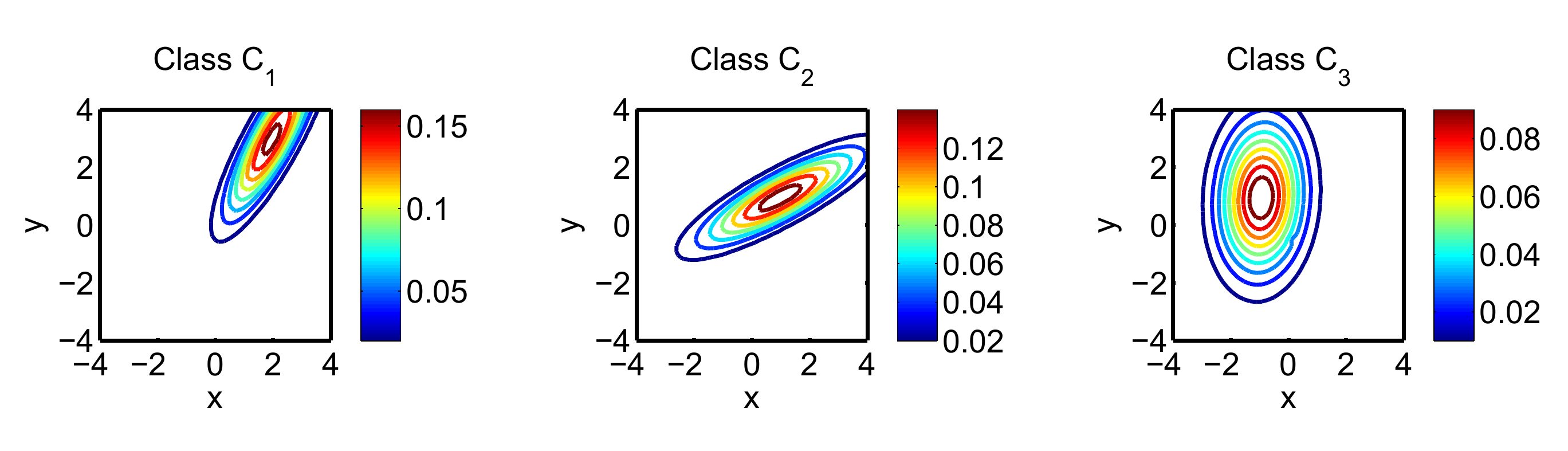}
	\caption{Iso-contour lines of data estimated with an $\alpha$-stable model.}
	\label{fig:est_stable}
\end{figure}

We extended our study to a vector of two dimensions to try to improve the classification rate.
\paragraph{The two dimension case}
~~\\
We generalized the study by considering a vector of two features. The learning base was used to estimate each mean and matrix of covariance for the Gaussian distributions (Figure~\ref{fig:est_gauss}) and each parameter $\alpha$, ${\boldsymbol\sigma}$, ${\boldsymbol\theta}$ and ${\boldsymbol\delta}$ for the $\alpha$-stable distributions~\cite{nolan2001estimation} (Figure~\ref{fig:est_stable}).
 
As in one dimension, the two null hypotheses $H_1$ and $H_2$ were verified (Table~\ref{test chi 2 2D generes}).\
The classification results (Table~\ref{comparaison_classif}) are somewhat less than impressive because there is confusion between classes (between $C_1$ and $C_2$ {\it cf}. Figure~\ref{fig:gen_gauss}). Furthermore, the work in two dimensions enables better results to be obtained than the combination. The Bayesian approach gives significantly the same results as the theory of belief functions. Indeed, the estimations of model and the estimation of prior probabilities are well-estimated.

We extended our study by considering synthetic data generated by $\alpha$-stable distributions.

\begin{table*}[htbp]    
\begin{center}
\begin{tabular}{|c|c|c|c|c|c|}
\cline{3-6}
\multicolumn{2}{c|}{}&\multicolumn{2}{|c|}{Bayesian approach}&\multicolumn{2}{|c|}{Belief approach} \\
\cline{3-6}
\multicolumn{2}{c|}{} & \scriptsize{Classification accuracies (\%)}&\scriptsize{95~\% confidence intervals}&\scriptsize{Classification accuracies (\%)}&\scriptsize{95~\% confidence intervals}\\
\hline
$H_1$& \scriptsize{1 dimension} & 66.19 &[65.13,67.25]&66.76&[65.71,67.81]\\
 \cline{2-6}
&\scriptsize{2 dimensions} & 76.14 &[75.19,77.09]&75.35&[74.39,76.31]\\
 \hline
$H_2$& \scriptsize{1 dimension} & 66.25 &[65.19,67.31]&67.15&[66.10,68.20]\\
 \cline{2-6}
&\scriptsize{2 dimensions} & 76.31 &[75.36,77.26]&74.50&[73.53,75,47]\\
 \hline 
\end{tabular}
\caption{Classification accuracies and 95~\% confidence intervals.}
\label{comparaison_classif}
\end{center}
\end{table*}

\begin{table*}[htbp]    
\begin{center}
\begin{tabular}{|c|c|c|c|c|} 
\cline{2-5}
\multicolumn{1}{c|}{} & \multicolumn{2}{|c|}{$H_1$}&\multicolumn{2}{|c|}{$H_2$}\\
\cline{2-5}
\multicolumn{1}{c|}{} & p-value&ksstat&p-value&ksstat\\
\hline
$C_1$& 0.1509& 0.0873&0.1509&0.0873\\
\hline
$C_1$& 0.0490& 0.0650&0.1997&0.0836\\
 \hline
 $C_3$& 0.1399& 0.0870&0.1665&0.0841\\
 \hline
\end{tabular}
\caption{Statistical values for a Kolmogorov-Smirnov test with a significance level of 5~\% (p-value: the critical value to reject the null hypothesis, ksstat: the greatest discrepancy between the observed and expected cumulative frequencies).}
\label{test chi 2 2D generes}
\end{center}
\end{table*}

\subsection{Application to synthetic data generated by $\alpha$-stable distributions}
In this subsection, we show the interest in modeling features in one and two dimensions with a single $\alpha$-stable distribution.
\subsubsection{Presentation of the data}
\begin{figure}[h]
	\centering
		\includegraphics[scale=0.53]{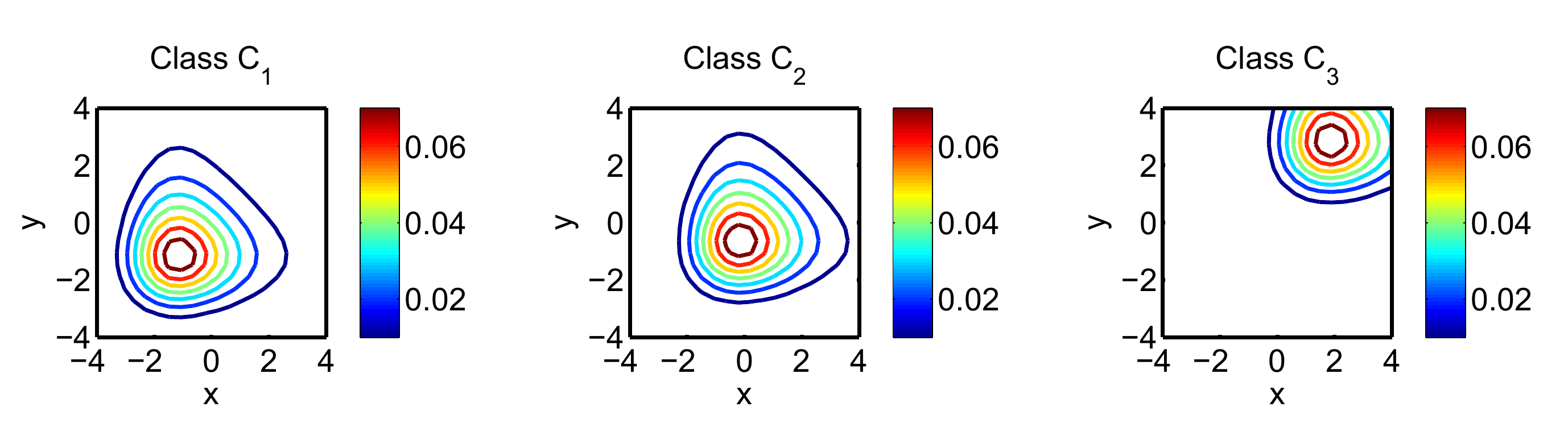}
\caption{Iso-contour lines of generated $\alpha$-stable pdf.}
\label{fig:comparaison_pdf_2D_généré}
\end{figure}

\begin{table*}[h!]    
\begin{center}
\begin{tabular}{|c|c|c|c|c|} 
\cline{2-5}
\multicolumn{1}{c|}{} & $\alpha$& ${\boldsymbol\sigma}$& ${\boldsymbol\theta}$& ${\boldsymbol\delta}$\\
\hline
First $\alpha$-stable {\it pdf} ($C_1$)& 1.5& $\displaystyle\begin{pmatrix}1 \\1\end{pmatrix}$  & $\begin{pmatrix}0 \\\frac{\pi}{2}\end{pmatrix}$& $\begin{pmatrix}0 \\0\end{pmatrix}$\\
\hline
Second $\alpha$-stable {\it pdf} ($C_2$)& 1.5& $\begin{pmatrix}1 \\1\end{pmatrix}$  & $\begin{pmatrix}0 \\\frac{\pi}{2}\end{pmatrix}$& $\begin{pmatrix}2 \\1.4\end{pmatrix}$\\
\hline
Third $\alpha$-stable {\it pdf} ($C_3$)& 1.5& $\begin{pmatrix}1 \\1\end{pmatrix}$  & $\begin{pmatrix}0 \\\frac{\pi}{2}\end{pmatrix}$& $\begin{pmatrix}1 \\0.5\end{pmatrix}$\\
\hline
\end{tabular}
\caption{Numerical values for the $\alpha$-stable {\it pdf}.}
\label{comparaison2}
\end{center}
\end{table*}
Three classes of artificial data sets were generated from 2D $\alpha$-stable distributions~\cite{modarres1994method} ({\it c.f.} Table~\ref{comparaison2} and \cite{nolan1995calculation} for the significance of ${\boldsymbol\sigma}$ and ${\boldsymbol\theta}$) with 3000 samples. A mesh between $[-4,4]\times [-4,4]$ enables the level curves of {\it pdf} to be plotted (Figure~\ref{fig:comparaison_pdf_2D_généré}).

\subsubsection{Results}

\begin{figure}[h]
	\centering
		\includegraphics[scale=0.4]{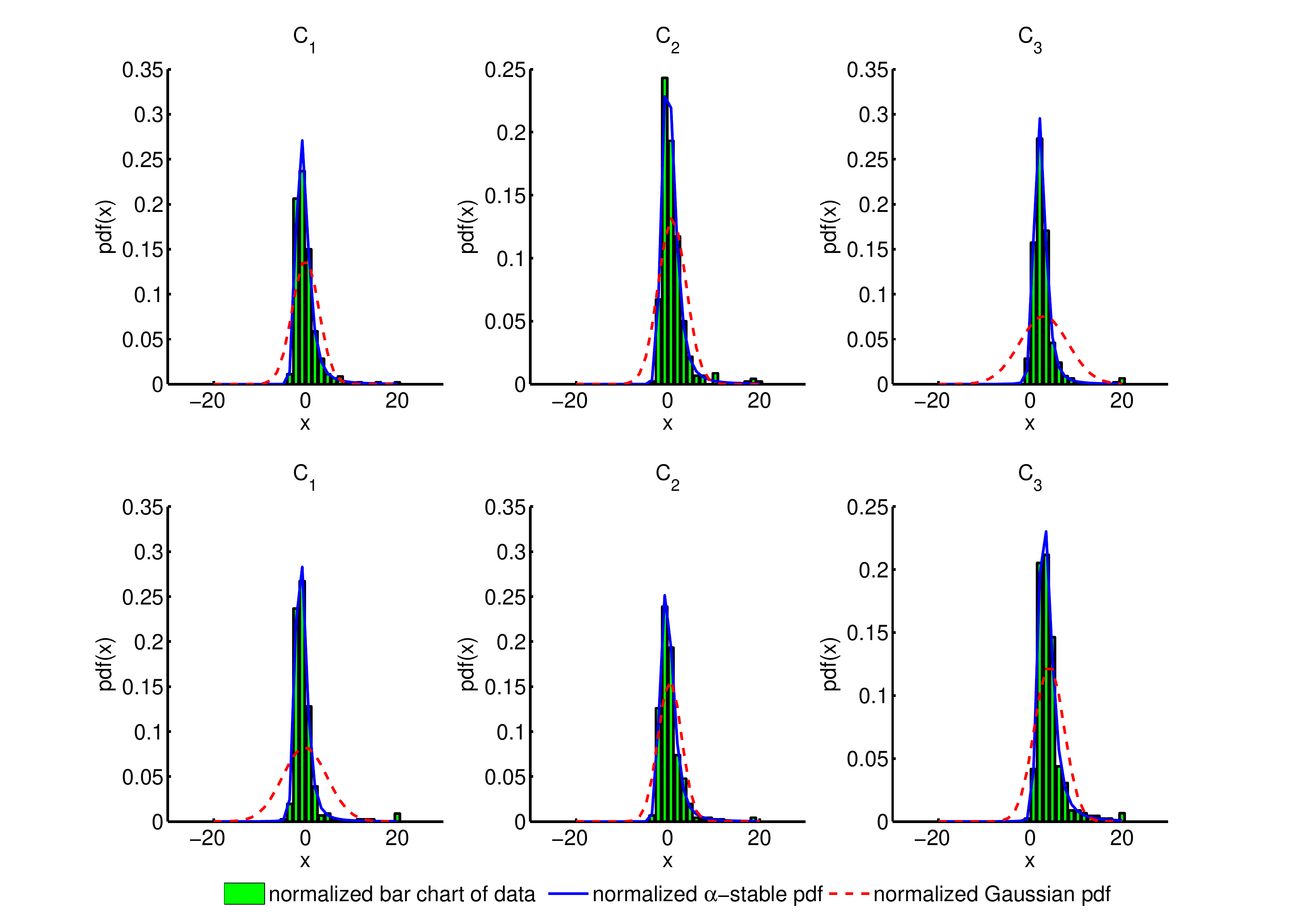}
	\caption{Empirical {\it pdf} and its estimations (The first row corresponds to the first feature and the second row corresponds to the second feature).}
	\label{fig:estimation_1D_generees}
\end{figure}

\paragraph{The one dimension case}
~~\\
We can assert that the estimation with the $\alpha$-stable hypothesis is better than the Gaussian hypothesis (Figure~\ref{fig:estimation_1D_generees}). There is a difference between the mode of the Gaussian model and the mode of $\alpha$-stable model. Indeed, the mean, calculated from samples, for the Gaussian model does not correspond to the mode when samples do not belong to a Gaussian law. Moreover, the values of {\it pdf}s are not the same compared to the real {\it pdf}s.   
\begin{table*}[htbp]    
\begin{center}
\begin{tabular}{|c|c|c|c|} 
\cline{3-4}
\multicolumn{2}{c|}{} & p-value&ksstat\\
\hline
$C_1$& first feature& 0.1758& 0.0851\\
\cline{2-4}
 & second feature& 0.7614 &0.0517\\
 \hline
 $C_2$& first feature& 0.8487& 0.0458\\
\cline{2-4}
 & second feature& 0.4794 &0.0630\\
 \hline
 $C_3$& first feature& 0.5505& 0.0621\\
\cline{2-4}
 & second feature& 0.2753 &0.0776\\
 \hline
\end{tabular}
\caption{Statistical values for a Kolmogorov-Smirnov test with a significance level of 5~\% (p-value: the critical value to reject the null hypothesis, ksstat: the greatest discrepancy between the observed and expected cumulative frequencies).}
\label{test chi 2 1D generes}
\end{center}
\end{table*}

\begin{figure}[h]
	\centering
		\includegraphics[scale=0.53]{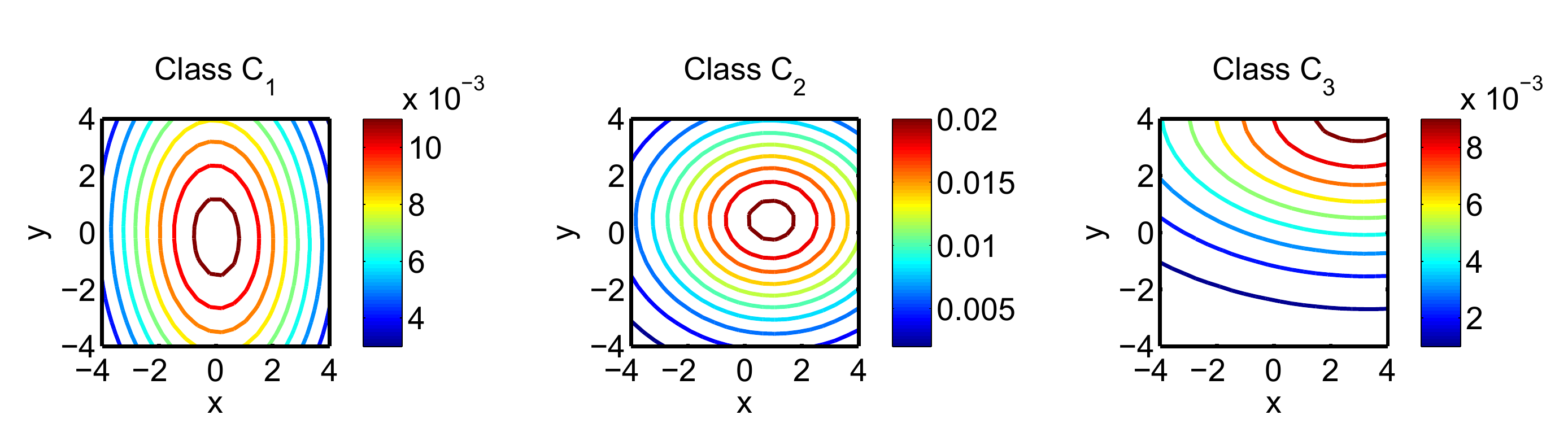}
	\caption{Iso-contour lines of data estimated with a Gaussian model.}
	\label{fig:estimation_Gaussien_générées_2D}
\end{figure}

\begin{figure}[h]
	\centering
		\includegraphics[scale=0.53]{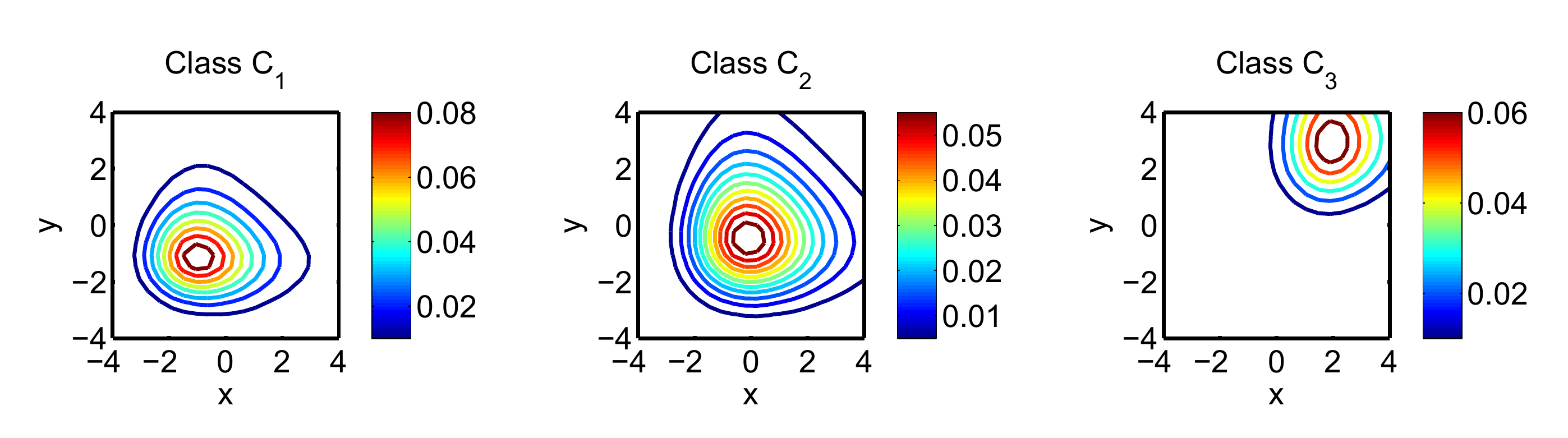}
	\caption{Iso-contour lines of data estimated with an $\alpha$-stable model.}
	\label{fig:alpha_stable_estime_2D}
\end{figure}

The K-S test is not satisfied for the null hypothesis $H_1$ for each class and each component. Consequently, we stopped the classification step for the Gaussian model. However, the K-S test is valid for the null hypothesis $H_2$ (Table~\ref{test chi 2 1D generes}). We can observe that the results obtained with the theory of belief functions are significantly the same as the results obtained using the Bayesian approach. This phenomenon can be explained by the fact that prior probabilities were well-estimated. Consequently, we changed voluntarily the values of prior probabilities. We fixed arbitrarily the prior probabilities: $p(C_1)=1/6$, $p(C_2)=2/3$ and $p(C_3)=1/6$. The classification accuracy decreased to 53.10~\% (Table~\ref{comparaisonprior}). Indeed, the learning base and the test base can sometimes be significantly different. For example, the prior probability of class sand can be greater than the prior probabilities of classes rock and silt, although it does not represent the truth, and the classification rate of class sand can be favoured. This phenomenon introduces confusion between classes and the classification rate can decrease. Consequently, the learning step is an important step for the Bayesian approach and can have a bad effect on the classification rate whereas the theory of belief functions is less sensitive to the learning step.   
\begin{table*}[htbp]    
\begin{center}
\begin{tabular}{|c|c|c|c|} 
\cline{3-4}
\multicolumn{2}{c|}{}&\multicolumn{2}{|c|}{Bayesian approach}\\
\cline{3-4}
\multicolumn{2}{c|}{} & \scriptsize{Classification accuracies (\%)}&\scriptsize{95~\% confidence intervals}\\
\hline
$H_2$& \scriptsize{1 dimension} & 55.26 &[54.15;56.37]\\
 \cline{2-4}
 & \scriptsize{2 dimensions} & 53.10 &[51.98;54.22]\\
 \hline
\end{tabular}
\caption{Classification accuracies and 95~\% confidence intervals with the prior probabilities $p(C_1)=1/6$, $p(C_2)=2/3$ and $p(C_3)=1/6$.}
\label{comparaisonprior}
\end{center}
\end{table*}
\paragraph{The two dimension case}
~~\\
We considered a vector of two features. In Figure~\ref{fig:estimation_Gaussien_générées_2D}, we draw the iso-contour lines of data estimated with the Gaussian model and in Figure~\ref{fig:alpha_stable_estime_2D} the iso-contour lines of data estimated with the $\alpha$-stable model \\
As in one dimension, the null hypothesis $H_1$ is rejected at significance level 5~\%, but the null hypothesis $H_2$ is valid (Table~\ref{test chi 2 2D generes alpha}).\\ 
The classification results (Table~\ref{comparaison33}) are somewhat less than impressive because there is confusion between classes. Furthermore, we obtain the same classification accuracies with the vector of two dimensions compared to the combination of two features. However, the classification decreases when the prior probabilities are not well-estimated (Table~\ref{comparaisonprior}). 

\begin{figure}[t]
	\centering
		\includegraphics[scale=0.35]{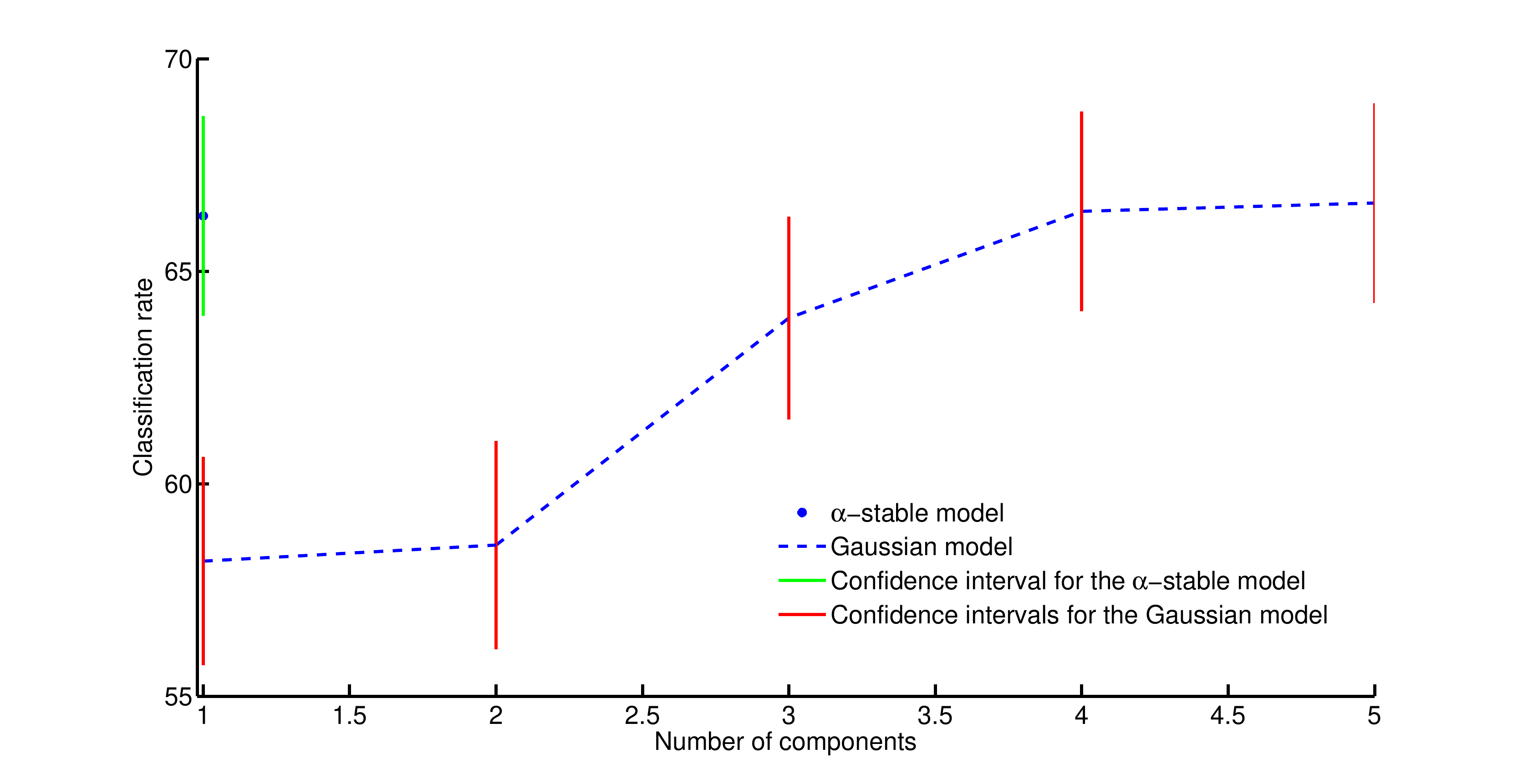}
	\caption{Classification rates according the number of components.}
	\label{fig:evolution_classification}
\end{figure}

We also compared these results with a GMM. It is well-known that the GMM can accurately model an arbitrary continuous distribution. This model estimation can be performed efficiently using the Expectation-Maximization (EM) algorithm. We compared classification accuracies by increasing the number of components. We drew the classification accuracy (Figure~\ref{fig:evolution_classification}) in relation to the number of components. Let us assume the null hypothesis:
\begin{itemize}
\item $G_n$: ``Data follow a Gaussian mixture model with $n$ components''.
\end{itemize}

\begin{figure}
	\centering
		\includegraphics[scale=0.6]{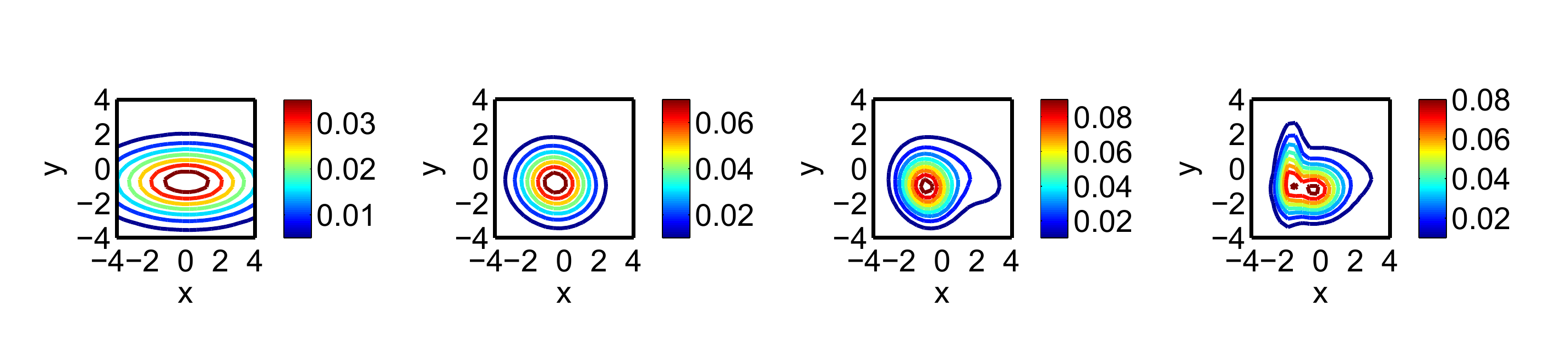}
		
	\caption{Representation of iso-contour by modeling data with a mixture of Gaussian for the class $C_1$ (from left to right: 2, 3, 4 and 5 components).}
	\label{fig:evolution_gmm_C1}
\end{figure}

\begin{figure}
	\centering
		\includegraphics[scale=0.6]{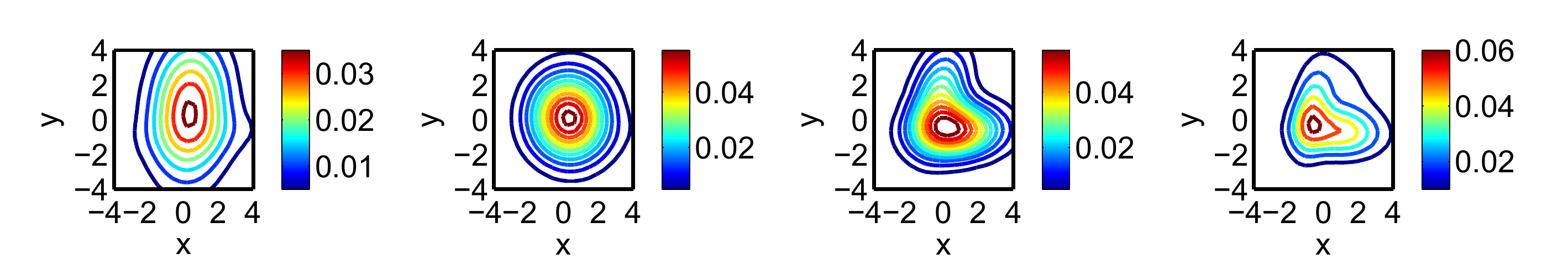}
	\caption{Representation of iso-contour by modeling data with a mixture of Gaussian for the class $C_2$ (from left to right: 2, 3, 4 and 5 components).}
	\label{fig:evolution_gmm_C2}
\end{figure}

\begin{figure}
	\centering
		\includegraphics[scale=0.6]{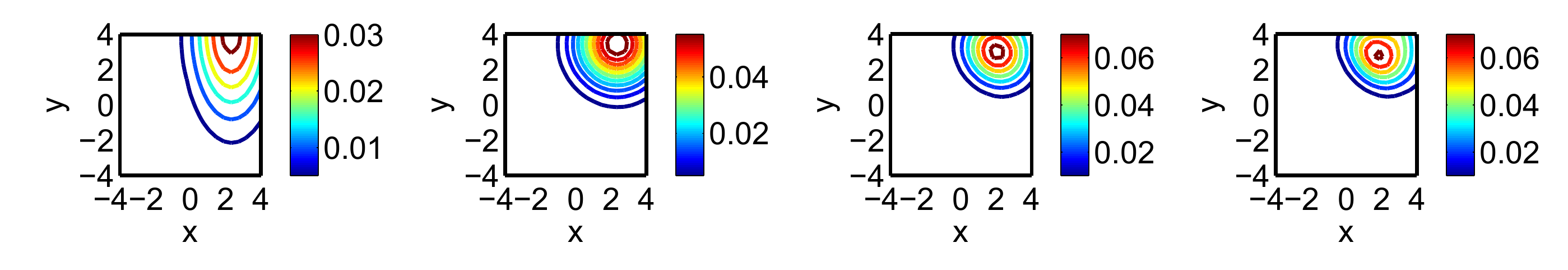}
	\caption{Representation of iso-contour by modeling data with a mixture of Gaussian for the class $C_3$ (from left to right: 2, 3, 4 and 5 components).}
	\label{fig:evolution_gmm_C3}
\end{figure}

The K-S test rejects the null hypothesis $G_n$ for a number of components $n<4$. The representations of iso-contour lines (Figure~\ref{fig:evolution_gmm_C1}, Figure~\ref{fig:evolution_gmm_C2} and Figure~\ref{fig:evolution_gmm_C3}) confirm the K-S test. Furthermore, the classification accuracies are lower than 
classification rates with a single $\alpha$-stable {\it pdf}. 

However, we can not increase indefinitely the number of components because a phenomenon of overfitting can appear (for example the representation of iso-contour by using a GMM with 5 components). Consequently, the number of 4 components seems to be a good compromise between the overfitting and good estimation. The classification accuracies are roughly the same between the single $\alpha$-stable model and the GMM with 4 components.   
\begin{table*}[htbp]    
\begin{center}
\begin{tabular}{|c|c|c|} 
\cline{2-3}
\multicolumn{1}{c|}{} & p-value&ksstat\\
\hline
$C_1$& 0.0683& 0.1003\\
\hline
$C_1$& 0.3694& 0.0688\\
 \hline
 $C_3$& 0.1154& 0.0932\\
 \hline
\end{tabular}
\caption{Statistical values for a Kolmogorov-Smirnov test with a significance level of 5~\% (p-value: the critical value to reject the null hypothesis, ksstat: the greatest discrepancy between the observed and expected cumulative frequencies).}
\label{test chi 2 2D generes alpha}
\end{center}
\end{table*}

\begin{table*}[htbp]    
\begin{center}
\begin{tabular}{|c|c|c|c|c|c|} 
\cline{3-6}
\multicolumn{2}{c|}{}&\multicolumn{2}{|c|}{Bayesian approach}&\multicolumn{2}{c|}{Belief approach}\\
\cline{3-6}
\multicolumn{2}{c|}{} & \scriptsize{Classification accuracies (\%)}&\scriptsize{95~\% confidence intervals}& \scriptsize{Classification accuracies (\%)}&\scriptsize{95~\% confidence intervals}\\
\hline
$H_2$& \scriptsize{1 dimension} & 66.56 &[65.51;67.62]&66.49&[65.44;67.55]\\
 \cline{2-6}
 & \scriptsize{2 dimensions} & 66.05 &[64.99;67.11]&65.16&[64.08;66.22]\\
 \hline
\end{tabular}
\caption{Classification accuracies and 95~\% confidence intervals.}
\label{comparaison33}
\end{center}
\end{table*}

\subsection{Application to real data from a mono-beam echo-sounder}

\subsubsection{Presentation of data}
\begin{figure}[b]
	\centering
		\includegraphics[height=6cm,width=12cm]{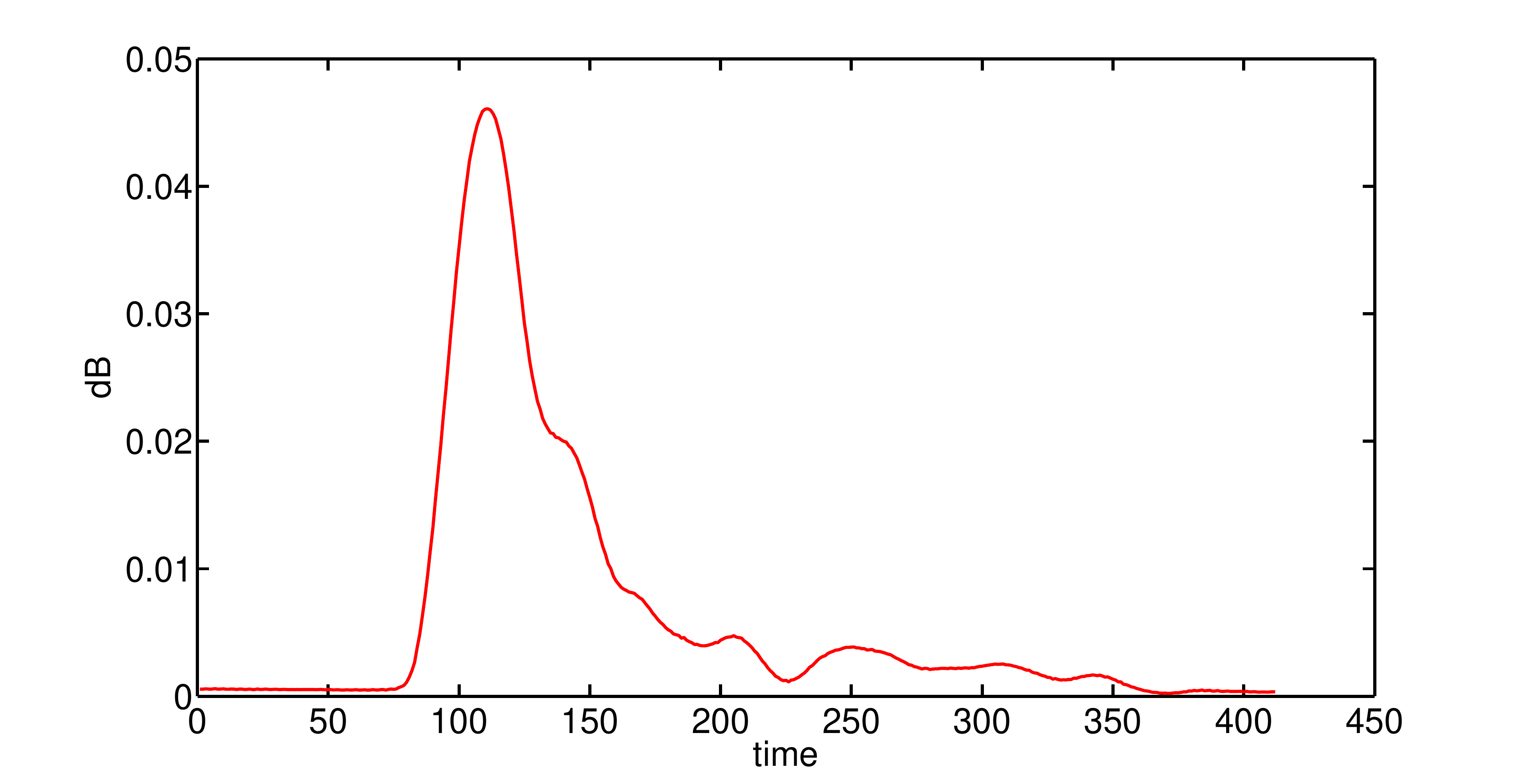}
	\caption{Example of echo signal amplitude from silt in low frequency.}
\label{fig:trace_signal_brute}
\end{figure}

\begin{figure}[h]
	\centering
		\includegraphics[height=6cm,width=12cm]{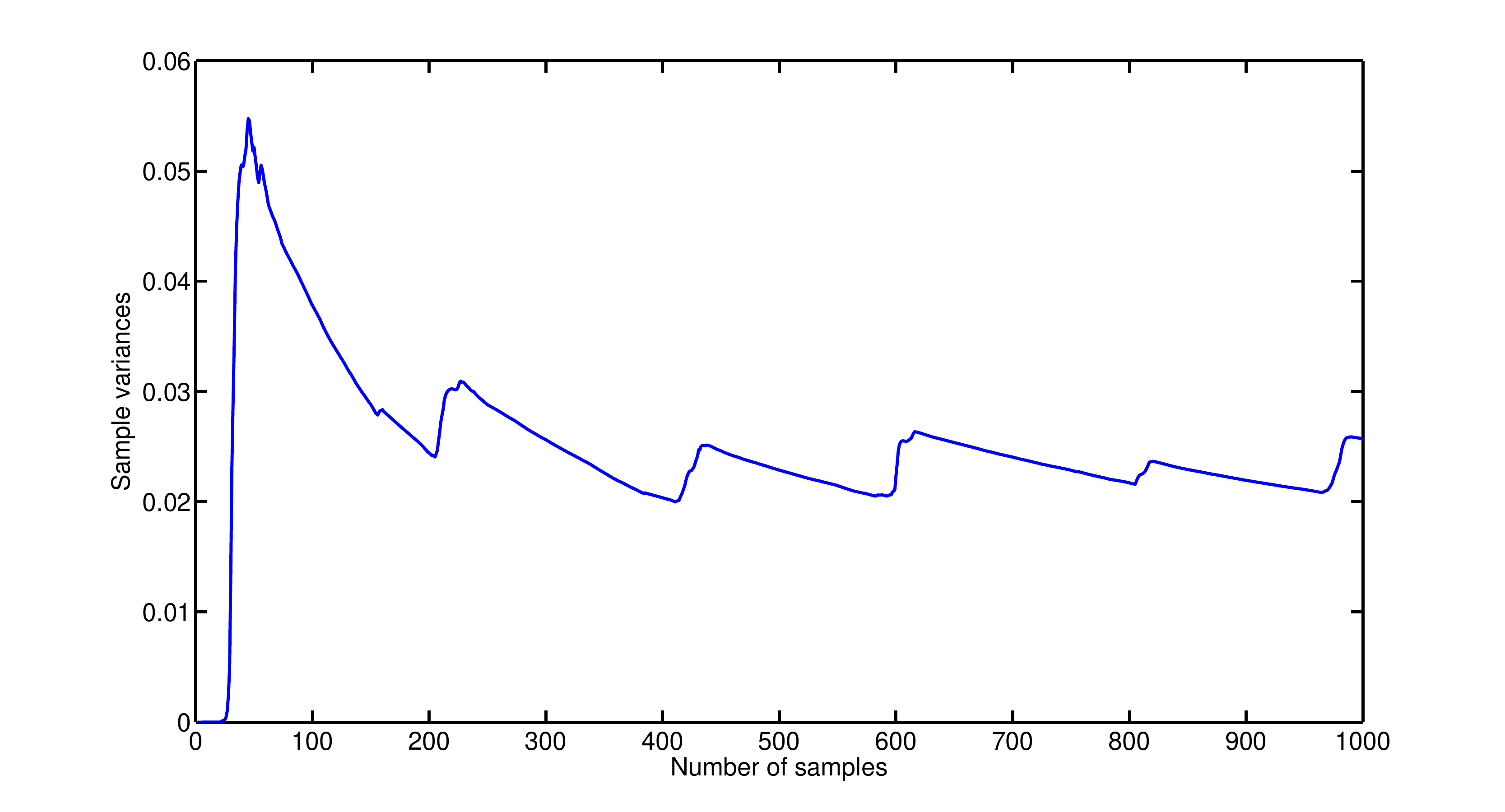}
	\caption{Running sample variances.}
		\label{fig:test_variance}
\end{figure}

We used a data set from a mono-beam echo-sounder given by the Service Hydrographique et Oc\'eanique de la Marine (SHOM). Raw data represents an echo signal amplitude according to time. In~\cite{Lurton}, the authors classify seven types of sea floor by comparing the time envelope of echo signal amplitude with a set of theoretical references curves. In our application, raw data (Figure~\ref{fig:trace_signal_brute}) were processed using the Quester Tangent Corporation (QTC) software~\cite{caughey1994sea} to obtain some features, which were normalized between [0,1]. These data had already been used for the navigation of an Autonomous Underwater Vehicle (AUV)~\cite{martinautonomous}. The frame of discernment is $\Theta=\{\text{silt},\text{rock},\text{sand}\}$, with 4853 samples from silt, 6017 from rock and 7338 from sand. The selection of features plays an important role in classification because the classification accuracy can be different according to the features. In~\cite{karoui}, the author gave an overview of methods based on features to classify data and choose features which discriminate all classes. Some features calculated by the QTC software can be modeled by an $\alpha$-stable {\it pdf}. There is a graphical test called converging variance test~\cite{granger1972infinite} which enables us to determine if samples belong to an $\alpha$-stable distribution. This test calculates the estimation of variances by varying the number of samples. If the estimation of variance diverges, the samples can belong to an $\alpha$-stable distribution. We give an example of this test for the feature called ``third quantile calculated on echo signal amplitude'' of the class rock (Figure~\ref{fig:test_variance}). We chose the features called  the ``third quantile calculated on echo signal amplitude'' and the ``25th quantile calculated on cumulative energy''.
\subsubsection{Results}
\begin{figure}[h]
	\centering
		\includegraphics[scale=0.4]{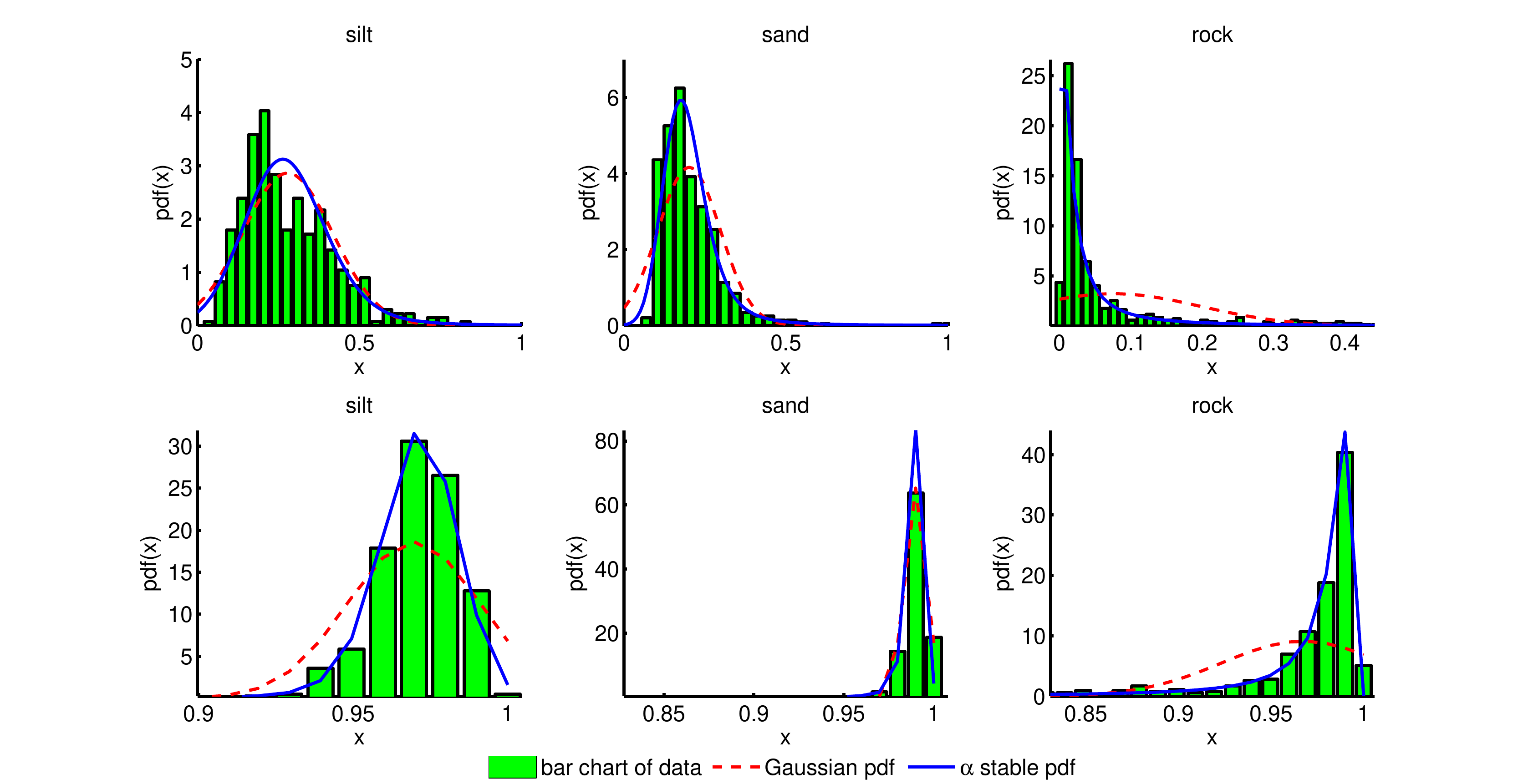}
	\caption{Empirical {\it pdf} and its estimations (The first row corresponds to the feature called ``third quantile calculated on echo signal amplitude'' and the second row corresponds to the feature called ``25th quantile calculated on cumulative energy'').}
	\label{fig:tracer_donnes}
\end{figure}

We randomly select 5000 samples for the data set. Half the samples were used for the learning base and the rest for the test base. We first consider features in one dimension, {\it i.e.} two mass functions were calculated and combined to obtain a single mass function.
\paragraph{The one dimension case} 
~~\\
We can observe that the assumption of the $\alpha$-stable model can easily accommodate the data compared to the Gaussian model (Figure~\ref{fig:tracer_donnes}). 

The $\alpha$-stable hypothesis is valid with the K-S test whereas the K-S test rejects the Gaussian hypothesis (Table~\ref{test chi 2 1D reelles}). The approach using the theory of belief functions (classification accuracy of 82.68~\%) give better results than the Bayesian approach (classification accuracy of 80.64~\%) but not significantly. This phenomenon can be explained by the fact that the estimated models and prior probabilities are well-estimated.

\begin{figure}[h]
	\centering
		\includegraphics[scale=0.54]{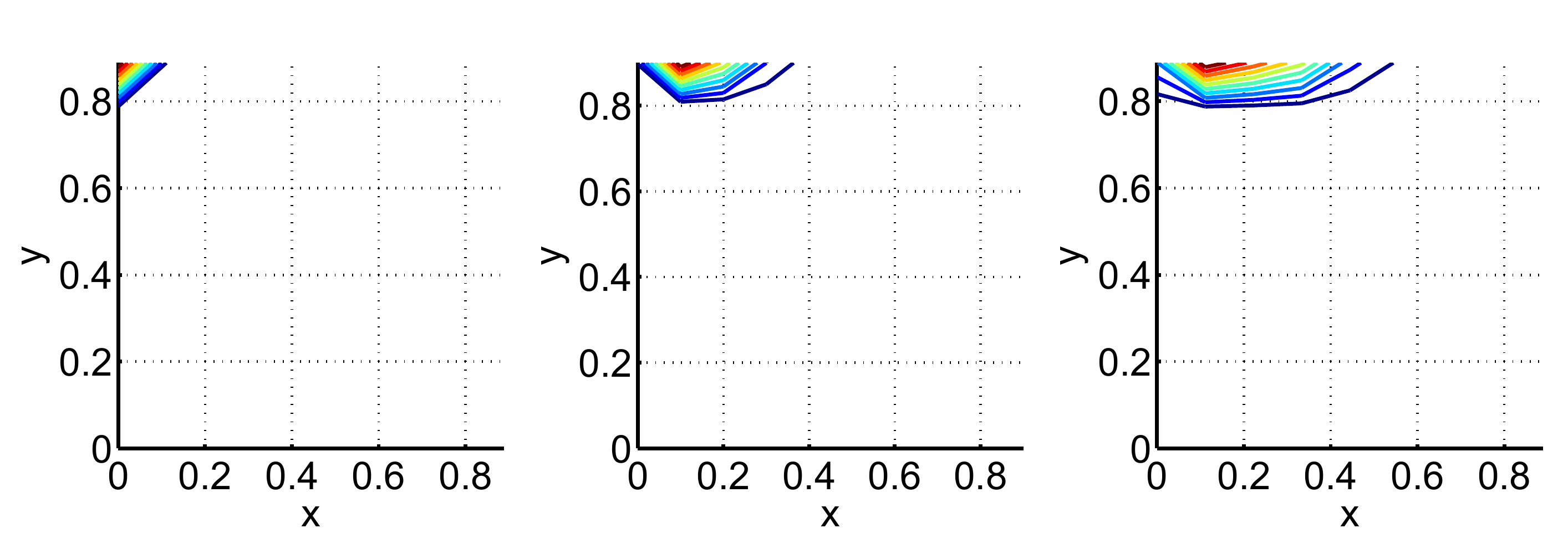}
	\caption{Level curves of three empirical $\alpha$-stable distributions with 10000 samples (The first column refers to the class rock, the second column refers to the class sand and the third column refers to the class silt).}
	\label{fig:donne_empiriques_reelles}
\end{figure}

\begin{figure}[t]
	\centering
		\includegraphics[scale=0.54]{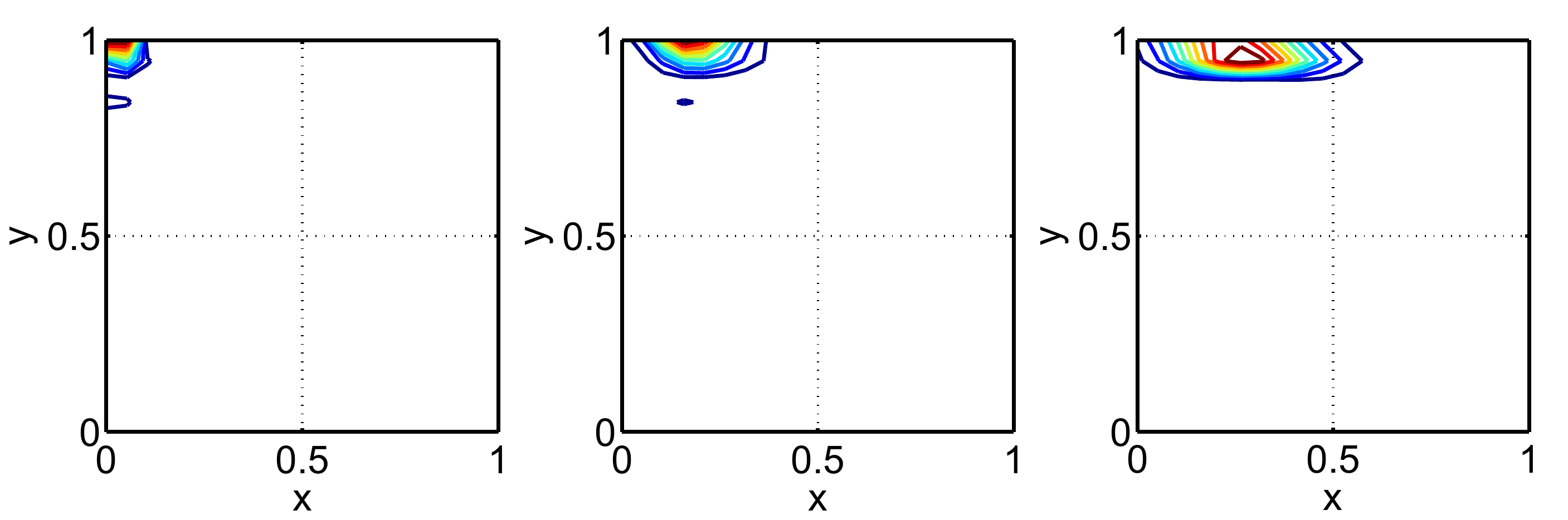}		
\caption{Level curves with an $\alpha$-stable model estimation (The first column refers to the class rock, the second column refers to the class sand and the third column refers to the class silt).}
\label{fig:estimer_reelles_alpha}
\end{figure}

\begin{figure}[h]
	\centering
		\includegraphics[scale=0.54]{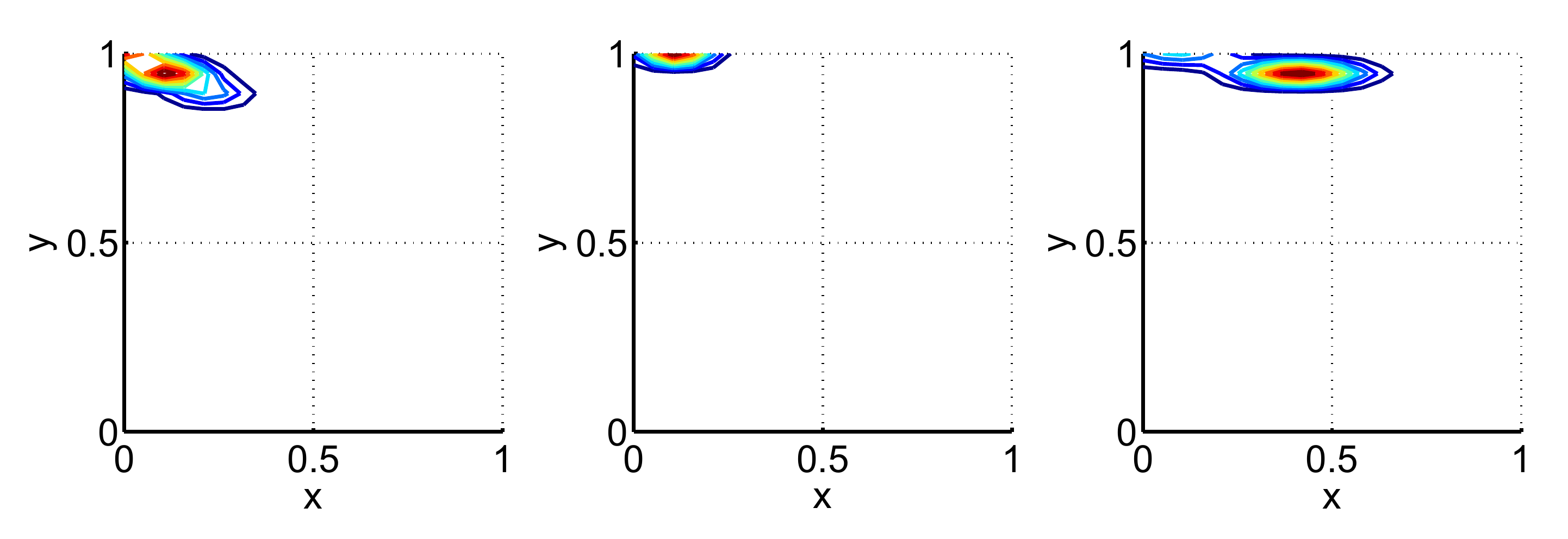}		
\caption{Level curves with a Gaussian model estimation (The first column refers to the class rock, the second column refers to the class sand and the third column refers to the class silt).}
\label{fig:donne_gaussienne_reelles}
\end{figure}

\paragraph{The two dimensions case} 
~~\\
We extended the study by using a vector of two features.  
The level curves of empirical data (Figure~\ref{fig:donne_empiriques_reelles}) show there is confusion between the class silt and the class sand. The $\alpha$-stable model (Figure~\ref{fig:estimer_reelles_alpha}) considers the confusion between classes compared to the Gaussian model (Figure~\ref{fig:donne_gaussienne_reelles}).

\begin{table*}[htbp]    
\begin{center}
\begin{tabular}{|c|c|c|c|} 
\cline{3-4}
\multicolumn{2}{c|}{} & p-value&ksstat\\
\hline
silt& first feature& 0.2310& 0.0879\\
\cline{2-4}
 & second feature& 0.7240 &0.0586\\
 \hline
 sand& first feature& 0.1206& 0.0811\\
\cline{2-4}
 & second feature& 0.1419 &0.0788\\
 \hline
 rock& first feature& 0.5784& 0.0595\\
\cline{2-4}
 & second feature& 0.0599 &0.1012\\
 \hline
\end{tabular}
\caption{Statistical values for a Kolmogorov-Smirnov test with a significance level of 5~\% (p-value: the critical value to reject the null hypothesis, ksstat: the greatest discrepancy between the observed and expected cumulative frequencies.)}
\label{test chi 2 1D reelles}
\end{center}
\end{table*}
\section{Conclusions}
In this paper, we have shown the advantages in using the $\alpha$-stable distributions to model data with heavy-tails. 
We considered the imprecision and uncertainty of data for modeling {\it pdf} and the theory of belief functions offers a way to model these constraints. We have examined the fundamental definitions of this theory and have shown a way to calculate plausibility functions when the knowledge of sensors is an $\alpha$-stable {\it pdf}. We finally validated our model by making a comparison with the model examined by Caron {\it et al.}~\cite{caron2006} where the {\it pdf} is a Gaussian.

Synthetic and real data were finally classified using the theory of belief functions. We estimated the learning base, calculated plausibility functions, combined plausibility functions and calculated maximum pignistic probability to obtain the classification accuracy. The K-S test shows that synthetic data can be modeled by an $\alpha$-stable model. The proposed method performs well for synthetic data. The consideration of two features improves the classification rate compared to the combination of two features for the Gaussian data. This study was then extended to a real application by using features extracted from a mono-beam echo-sounder. The Gaussian model is limited when modeling features compared to the $\alpha$-stable model because the $\alpha$-stable model has more degrees of freedom. To sum up, the interest in using an $\alpha$-stable model is that data, in both one or two dimensions, can be modeled. The problem with the $\alpha$-stable model is the computation time. Indeed, we proceed numerically to calculate plausibility functions (discretization for the bivariate $\alpha$-stable {\it pdf}). Furthermore, it is difficult to generalize in dimension $d>2$ without increasing CPU time.

We present preliminary results about classification problem by modeling features with $\alpha$-stable probability density functions in dimension $d\leq2$. In future works, we plan to make experiments in the high-dimensional case with $d\geq3$. One solution in perspective to improve the computation time is to optimize the code in another language such that C/C++. Another perspective to this work is the use of other features from sensors to improve the classification accuracy, images from side-scan sonar.
\section*{Acknowledgements}
The authors would like to thank the Service Hydrographique et Oc\'eanique de la Marine (SHOM) for the data and G. Le Chenadec for his advice concerning the data.
\bibliographystyle{model1a-num-names}

\clearpage

\clearpage

\end{document}